\documentclass[11pt, a4paper, logo, copyright]{googledeepmind}

\usepackage[authoryear, sort&compress, round]{natbib}
\usepackage{times}
\usepackage{latexsym}
\usepackage{lineno}
\usepackage{subcaption}
\usepackage[utf8]{inputenc} % allow utf-8 input
\usepackage[T1]{fontenc}    % use 8-bit T1 fonts
\usepackage{hyperref}       % hyperlinks
\usepackage{url}            % simple URL typesetting
\usepackage{booktabs}       % professional-quality tables
\usepackage{amsfonts}       % blackboard math symbols
\usepackage{nicefrac}       % compact symbols for 1/2, etc.
\usepackage{microtype}      % microtypography
\usepackage[table,dvipsnames]{xcolor}         % colors
\usepackage{xspace}
\usepackage{pifont}
\usepackage{booktabs}
\usepackage{multirow}
\usepackage{graphicx}
\usepackage{wrapfig}
\usepackage{makecell}
\usepackage{xspace}
\usepackage{comment}
\usepackage[disable]{todonotes}
\usepackage{enumitem}
\usepackage{amsmath}
\usepackage{titlesec}
\usepackage[most]{tcolorbox}

\usepackage{tcolorbox}
\newtcolorbox{prompt}[1]{colback=gray!20,colframe=gray!50!black,fonttitle=\bfseries,title=#1}

\newcommand{\method}{Evo-Memory\xspace}
\newcommand{\alg}{ReMem\xspace}

\title{Evo-Memory: Benchmarking LLM Agent Test-time Learning with Self-Evolving Memory}

\correspondingauthor{twei10@illinois.edu}

\reportnumber{} % Leave blank if n/a

\author[$\dagger$,1]{Tianxin Wei}
\author[2]{Noveen Sachdeva}
\author[2]{Benjamin Coleman}
\author[2]{Zhankui He}
\author[1]{Yuanchen Bei}
\author[1]{Xuying Ning}
\author[1]{Mengting Ai}
\author[$\dagger$,1]{Yunzhe Li}
\author[1]{Jingrui He}
\author[2]{Ed H. Chi}
\author[2]{Chi Wang}
\author[2]{Shuo Chen}
\author[2]{Fernando Pereira}
\author[2]{Wang-Cheng Kang}
\author[2]{Derek Zhiyuan Cheng}

\affil[$\dagger$]{Work done while at Google DeepMind}
\affil[1]{University of Illinois Urbana-Champaign}
\affil[2]{Google DeepMind}

\begin{abstract}
Statefulness is essential for large language model (LLM) agents to perform long-term planning and problem-solving. This makes \emph{memory} a critical component, yet its management and evolution remain largely underexplored. Existing evaluations mostly focus on static conversational settings, where memory is passively retrieved from dialogue to answer queries, overlooking the dynamic ability to accumulate and reuse \emph{experience} across evolving task streams. In real-world environments such as interactive problem assistants or embodied agents, LLMs are required to handle continuous task streams, yet often fail to learn from accumulated interactions, losing valuable contextual insights, a limitation that calls for \emph{test-time evolution}, where LLMs retrieve, integrate, and update memory continuously during deployment. To bridge this gap, we introduce \textbf{\method}, a comprehensive streaming benchmark and framework for evaluating \emph{self-evolving memory} in LLM agents. \method structures datasets into sequential task streams, requiring LLMs to search, adapt, and evolve memory after each interaction. We unify and implement over ten representative memory modules and evaluate them across 10 diverse multi-turn goal-oriented and single-turn reasoning and QA datasets. To better benchmark experience reuse, we provide a baseline method, \textbf{ExpRAG}, for retrieving and utilizing prior experience, and further propose \textbf{\alg}, an \emph{action–think–memory refine} pipeline that tightly integrates reasoning, task actions, and memory updates to achieve continual improvement.

\end{abstract}

\begin{document}

\maketitle

\addtocontents{toc}{\protect\setcounter{tocdepth}{-1}}

\section{Introduction}

Large Language Models (LLMs) have rapidly evolved from simple chatbots into capable systems that can write code, control browsers, and perform advanced question answering~\citep{comanici2025gemini}. These advances have been driven by improving inference, planning, and tool use, as shown by benchmarks emphasizing logical reasoning and multi-step actions. Yet a fundamental capability, \emph{memory}, remains largely underexplored. Memory allows LLMs to maintain state across interactions, accumulate experience, and adapt strategies over time. Recent studies have introduced memory modules that track dialogue histories through compression, indexing, or retrieval~\citep{maharana2024evaluating}, improving \emph{conversational recall} and personalization. However, most of these systems only reuse static dialogue context rather than learning from experience to improve future reasoning or decision-making.

Despite these advances, existing LLM memory systems remain largely static, retrieving information passively rather than evolving through use. Current evaluations test whether models can recall past context but rarely assess their ability to \emph{reuse experience}. In essence, agents remember what was said but not what was learned. \emph{Conversational recall} retrieves prior facts, whereas \emph{experience reuse} abstracts reasoning strategies for future tasks. Without such reuse, models repeatedly solve similar problems, as long-term assistants often recall context yet fail to adapt across sessions.

\begin{figure}[t]
    \centering
    \begin{subfigure}[t]{0.45\textwidth}
        \centering
        \includegraphics[width=\linewidth]{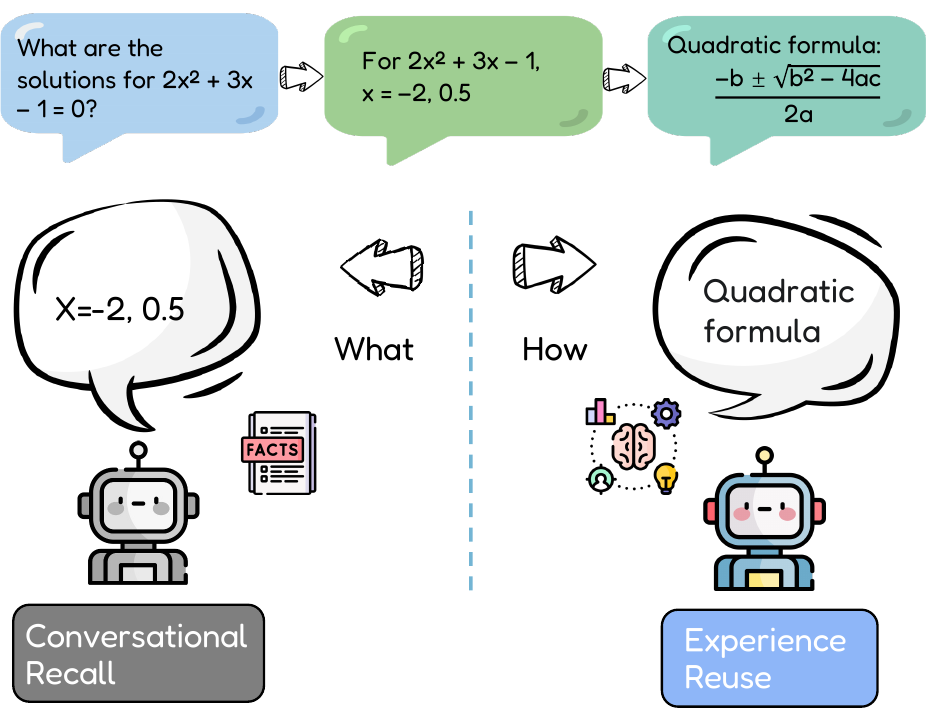}
        \caption{Recall vs. reuse.}
        \label{fig:compare}
    \end{subfigure}
    \hfill
    \begin{subfigure}[t]{0.52\textwidth}
        \centering
        \includegraphics[width=\linewidth]{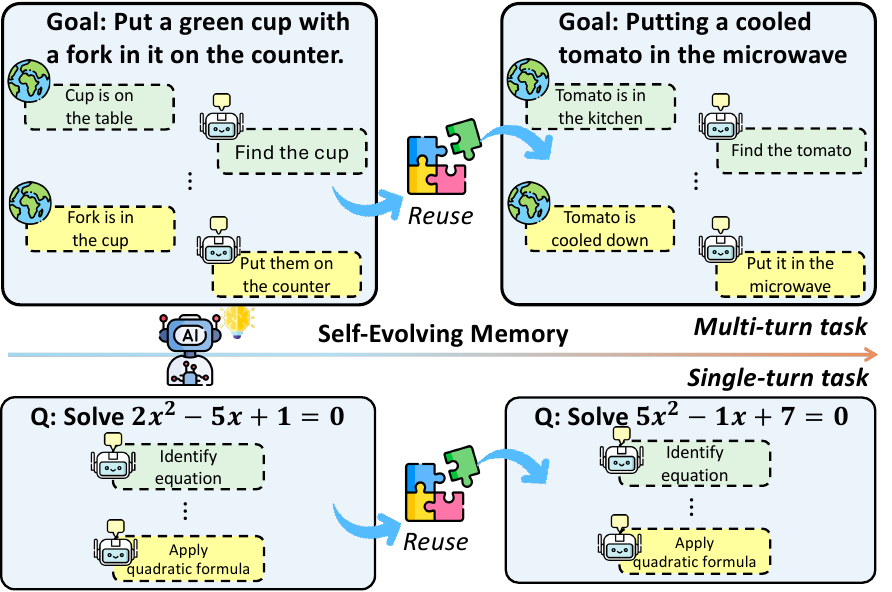}
        \caption{Task settings.}
        \label{fig:compare_task}
    \end{subfigure}
    \caption{\textbf{(a)} Conversational recall retrieves past facts (e.g., solutions to $2x^2 + 3x - 1 = 0$), while experience reuse recalls reasoning strategies (e.g., using the formula). \textbf{(b)} A stateful agent encounters both multi-turn tasks (e.g., embodied manipulation) and single-turn tasks (e.g., solving equations), and should learn reusable experiences from past interactions.}
    \label{fig:combined_compare}
    \vspace{-0.5cm}
\end{figure}

Several recent benchmarks have begun examining static adaptation but remain limited in scope. StreamBench~\citep{wu2024streambench} evaluates sequential learning but mainly measures factual retention without reasoning or trajectory reuse. LifelongBench~\citep{zheng2025lifelongagentbench} studies lifelong learning across environments and skills but focuses on retention without modeling memory structure or updates. Other studies~\citep{hu2025evaluating,wulongmemeval,maharana2024evaluating} assess long-term conversational consistency but do not test how agents evolve their memory during deployment. Together, these efforts highlight a critical gap: while progress has been made on sequential reasoning, there is still no unified framework for evaluating how different memory methods retrieve, integrate, and evolve historical strategies in realistic streaming scenarios. Figure~\ref{fig:compare} illustrates this contrast between static recall and cumulative improvement through self-evolving memory.

% \begin{figure}[t]
%     \centering
%     \includegraphics[width=0.43\textwidth]{figures/compare_figure.pdf}
%     \caption{Conversational recall retrieves past facts (e.g., solutions to $2x^2 + 3x - 1 = 0$).
%     Experience reuse recalls reasoning strategies (e.g., using the formula).}
%     \label{fig:compare}
% \end{figure}

To bridge this gap, we introduce \textbf{\method}, a comprehensive streaming benchmark and framework for evaluating \emph{self-evolving memory} in LLM agents. Figure~\ref{fig:compare_task} illustrates how a self-evolving agent reuses prior experiences across both multi-turn interactive tasks and single-turn reasoning tasks. \method restructures datasets into sequential \emph{task streams}, requiring models to retrieve, adapt, and evolve memory after each interaction. The benchmark covers both \emph{multi-turn goal-oriented} environments and \emph{single-turn reasoning or problem-solving} tasks, explicitly testing whether LLMs can accumulate knowledge and refine strategies during deployment, a process we term \emph{test-time evolution}. We unify and implement over ten representative memory modules, including retrieval-based, workflow, and hierarchical memory systems, to study their adaptation behavior. To further examine experience reuse, we introduce \textbf{ExpRAG}, a simple retrieval-based baseline that leverages prior task experiences, and further develop \textbf{\alg}, an advanced \emph{action–think–memory refine} pipeline that tightly integrates reasoning, action, and memory updates for continual improvement.
% \begin{figure}[t]
%     \centering
%     \includegraphics[width=1.05\linewidth]{figures/teaser.pdf}
%     \caption{Illustration of different task types and experience reusing. A stateful agent encounters both multi-turn tasks (e.g., embodied manipulation) and single-turn tasks (e.g., solving equations), and should learn reusable experiences from past experiences.}
%     \label{fig:compare_task}
% \end{figure}

In summary, our contributions are threefold:
\begin{itemize}[itemsep=0pt,topsep=2pt,leftmargin=*]
    \item \textbf{Benchmark:} We present \method, a streaming benchmark that evaluates LLM agents' ability to perform \emph{test-time evolution} across diverse multi-turn and single-turn tasks, bridging the gap between conversational recall and experience reuse.
    \item \textbf{Framework:} We provide a unified evaluation framework with memory-centric metrics for analyzing adaptation, efficiency, and stability, and will release all code and configurations for reproducibility.
\item \textbf{Analysis and Insights:} We introduce \textbf{ExpRAG}, a simple retrieval-based baseline for experience reuse, and \textbf{\alg}, an \emph{action–think–memory refine} pipeline that unifies reasoning, action, and memory for continual improvement, informing future designs of memory.
\end{itemize}

\section{Related Work}

In this section, we review existing works on test-time learning and self-evolving memory. 

\subsection{Test-time Learning}
Test-time learning (TTL) builds upon early work on test-time adaptation (TTA)~\citep{wang2021tent,zhang2023memo}, which enables models to adjust to distribution shifts during deployment. Recent advances extend TTA toward \emph{continuous self-improvement}~\citep{iwasawa2021t3a,liu2021ttt++}, allowing models to refine their behavior through online optimization.
Recent \emph{agent-based} studies operationalize such continual improvement via reflection, planning, and self-evolution. Works like \citep{shinn2023reflexion,wang2023voyager,ma2024eureka,park2023generative,he2025evotest,he2025enabling} and newer frameworks, including \citep{yang2024large} demonstrate how agents autonomously revise plans, synthesize feedback, and co-evolve \citep{gao2025survey}. These advances mark a shift from static adaptation toward adaptive, self-improving agents capable of continual learning during deployment.

\subsection{Self-evolving Memory}
Early LLM memory systems primarily served as \emph{passive storage}, maintaining recent dialogues or retrieved facts to compensate for limited context windows~\citep{DBLP:conf/nips/LewisPPPKGKLYR020,DBLP:conf/iclr/AsaiWWSH24,Zhong2023MemoryBankEL,Liu_LlamaIndex_2022,Packer2023MemGPTTL}. Subsequent studies introduced richer management mechanisms, including differentiable read–write controllers~\citep{Modarressi2023RETLLMTA,Liang2023SCMEL} and evaluations under realistic conversational settings~\citep{maharana2024evaluating,wulongmemeval}. Beyond static buffers, recent work explores \emph{policy-driven control}, where the model is explicitly optimized to decide what to store, retrieve, or overwrite~\citep{yu2025memagent,xu2025mem,zhou2025mem1,yan2025memory,Li2025MemOSAO}. Meanwhile, structured memory representations have emerged to organize experiences into relational or procedural forms, as in RepoGraph~\citep{Ouyang2024RepoGraphEA}, MEM0~\citep{mem0}, Zep~\citep{Rasmussen2025ZepAT}, and Dynamic Cheatsheets~\citep{suzgun2025dynamic}. However, there remains no unified evaluation setting and framework for \emph{self-evolving memory}, the ability to reuse and adapt experiences across tasks. \method builds on this trajectory by benchmarking how LLMs not only store and recall but also evolve, reorganize, and reuse memory under streaming task settings.

\section{\protect\method: Evaluating Self-Evolving Memory in LLM Agents}
\label{sec:method}

\begin{figure*}[t]
    \centering
    \includegraphics[width=\linewidth]{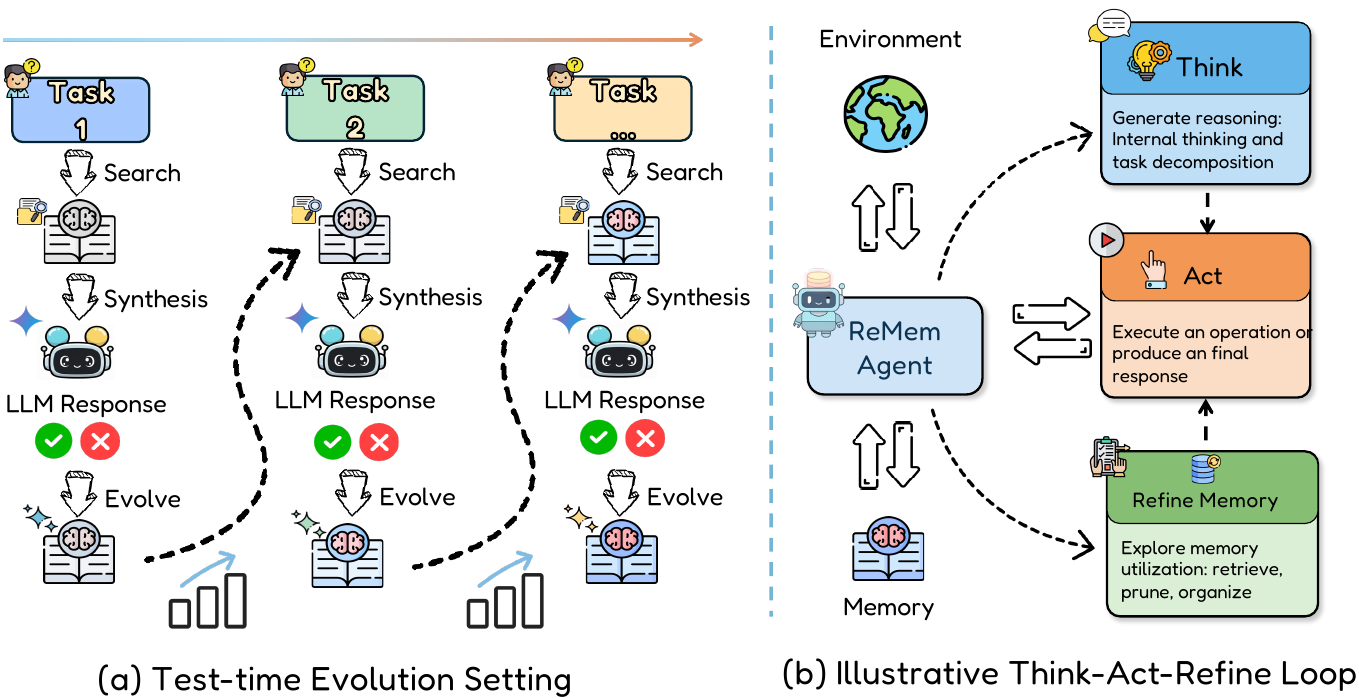}
    \caption{Overview of the ReMem agent framework. Left: Test-time evolution process where the agent iteratively searches, synthesizes, and evolves its memory across multiple tasks. Right: Agent architecture with three core modules, Think (reasoning and decomposition), Refine Memory (retrieve, prune, organize), and Act (execution), that interact with the environment and learned memory.}
    \label{fig:main}
\end{figure*}

Existing evaluations of LLMs often treat memory as static recall, overlooking its role in continual adaptation. 
\method provides a unified benchmark to study \emph{self-evolving memory}, where agents retrieve, integrate, and update knowledge over time. 
As illustrated in Figure~\ref{fig:main}, the left side shows the test-time evolution process, and the right side outlines the \textsc{ReMem} agent with three modules: \emph{Think}, \emph{Act}, and \emph{Refine Memory}. 
We first formalize the problem setting, then describe two representative implementations, \textsc{ExpRAG} and \textsc{ReMem}, used to instantiate the benchmark.

\subsection{Problem Formulation}

% We formalize a general memory-augmented agent as a pair $(F, U)$, where $F$ is the base LLM and $U$ is the memory update pipeline. At time $t$, the agent observes input $x_t$, maintains memory $M_t$, and produces an output $\hat{y}_t$. This abstraction unifies a wide spectrum of existing memory mechanisms, from retrieval-augmented generation to dynamic, hierarchical, and workflow-based memories, under a single iterative formulation.

We formalize a general memory-augmented agent as a tuple
(\text{F}, \text{U}, \text{R}, \text{C}), 
where \(\text{F}\) is the base LLM, \(\text{U}\) is the memory update pipeline, \(\text{R}\) is the retrieval module, and \(\text{C}\) is the contextual construction mechanism that transforms retrieved content into the final working context. In our setting, the agent processes a sequence of inputs \(\{x_1, x_2, \ldots, x_T\}\), and the memory state \(M_t\) evolves with the history. At time \(t\), the agent receives an input \(x_t\), maintains an evolving memory \(M_t\), retrieves relevant elements \(\text{R}(M_t, x_t)\), constructs a contextualized prompt 
\[
C_t = \text{C}(x_t, \text{R}(M_t, x_t)),
\]
and produces an output 
\[
\hat{y}_t = \text{F}(C_t).
\]
This abstraction unifies a wide spectrum of existing memory mechanisms, from retrieval-augmented generation to dynamic, hierarchical, and workflow-based memories, under a single iterative formulation.

\paragraph{Search.}  
Given the current input $x_t$, the agent first retrieves relevant memory entries:  
\[
R_t = \text{R}(M_t, x_t),
\]
where $\text{R}$ can represent similarity search, index-based lookup, or attention over stored embeddings. This step captures memory access policies across different algorithms.

% \paragraph{Synthesis.}  
% The agent interprets and restructures the retrieved information $R_t$ into a concise working context aligned with the current input $x_t$.  
% This synthesis yields a coherent text $\tilde{C}_t$, from which the final output is derived:
% \[
% \hat{y}_t = \text{F}(\tilde{C}_t).
% \]

\paragraph{Synthesis.}
The agent restructures the retrieved information \(R_t\) into a working context tailored to the current input \(x_t\).  
This step may involve forming a structured prompt \citep{Wang2024AgentWM}, selecting key memory items \citep{mem0,xu2025mem}, or merging retrieved content \citep{suzgun2025dynamic} into a short summary.  
We denote the resulting context as \(\tilde{C}_t = \text{C}(x_t, R_t)\), and the final output is
\[
\hat{y}_t = \text{F}(\tilde{C}_t).
\]

\paragraph{Evolve.}  

After obtaining \(\hat{y}_t\), the agent constructs a new memory entry
\(m_t = h(x_t, \hat{y}_t, f_t)\) that captures the current step’s experience together with the feedback \(f_t\), such as whether the task was completed. The memory is then updated via:
\[
M_{t+1} = \text{U}(M_t, m_t).
\]
Different algorithms instantiate $U$ differently, for example, direct append for retrieval-based memories, summarization or compression for long-term storage, or replacement for bounded-capacity stores.  
This unified formulation abstracts the essential cycle of \emph{retrieval}, \emph{synthesis}, and \emph{evolution} underlying all memory-based agents.

\paragraph{Dataset Preparation.}  
\method restructures conventional static datasets into \emph{streaming task sequences}, enabling evaluation of how LLMs reuse and evolve memory over time.  
Each dataset can thus be transformed into a sequence $\tau = \{(x_1, y_1), \dots, (x_T, y_T)\}$, forming a ground-truth trajectory in which earlier tasks provide essential information or strategies for later ones.  
At each step $t$, the agent processes input $x_t$, retrieves and synthesizes memory, produces prediction $\hat{y}_t$, and updates the memory state $M_t$, yielding the predicted trajectory:
\[
\scalebox{0.9}{$
(x_1, \hat{y}_1, M_1) \rightarrow (x_2, \hat{y}_2, M_2) \rightarrow \cdots \rightarrow (x_T, \hat{y}_T, M_T).
$}
\]
This design transforms static benchmarks into interactive evaluation streams that explicitly probe an LLM’s ability to accumulate, adapt, and refine knowledge during deployment.

\subsection{\protect\textbf{ExpRAG}: Experience Retrieval and Aggregation}

As a simple baseline and extension, we define \textbf{ExpRAG}, a task-level retrieval-augmented agent. Each memory entry $m_i = S(x_i, \hat{y}_i, f_i)$ encodes a structured experience text with template $S$.
At step $t$, the agent retrieves $k$ similar experiences from memory according to a retrieval score $\phi$:
\[
R_t = \text{Top-}k_{m_i \in M_t}~\phi(x_t, m_i).
\]
The model conditions on these retrieved examples following the in-context learning principle:
\[
\hat{y}_t = \text{F}(x_t, R_t),
\]
and appends the new experience to memory:
\[
M_{t+1} = M_t \cup \{(x_t, \hat{y}_t, f_t)\}.
\]
ExpRAG thus performs one-shot experience reuse through retrieval and aggregation. It captures how simple memory-based extensions of in-context learning behave but lacks iterative reasoning or adaptive refinement during inference.

\subsection{\protect\alg: Synergizing Reasoning, Acting, and Memory}

We propose \textbf{\alg}, a simple yet effective framework that unifies reasoning, action, and memory refinement within a single decision loop.
Unlike conventional retrieval-augmented or ReAct-style methods that treat memory as static context, \alg introduces a third dimension of \emph{memory reasoning}, allowing the agent to actively evaluate, reorganize, and evolve its own memory during problem solving.

At each step $t$, given the current input $x_t$, memory state $M_t$, and previous reasoning traces $o^{1:n-1}_t$ at this step,  the agent selects one of the operations:
\[
a_t^n \in \{\text{Think},~\text{Act},~\text{Refine}\}.
\]
It then performs the operation and transitions according to:
\[
o_t^n = \text{Agent}(x_t, M_t, a_t^n),
\]
where $o_t^n$ denotes the output generated at step $t$ after $n$ operations, such as an intermediate reasoning trace, an external action, and memory refine thoughts.

Specifically,  
\emph{Think} produces internal reasoning traces that help decompose the task and guide subsequent actions;  
\emph{Act} executes an operation in the environment or outputs a response observable to the user;  
\emph{Refine} performs meta-reasoning over memory, which exploiting useful experiences, pruning noise, and reorganizing $M_t$, to better support future reasoning and action. Within each step, the agent may perform multiple rounds of \emph{Think} and \emph{Refine}, and the step terminates once an \emph{Act} operation is selected.  
This induces a Markov decision process where the state at step \(t\) after $n$ operations is \(s_t^n = (x_t, M_t, o^{1:n-1}_t)\), the action space is \(\{\text{Think}, \text{Act}, \text{Refine}\}\), and the transition dynamics are given by the \(\text{Agent}\) operator together with the environment response.  
Depending on the task, the \textit{Act}-output of step \(t\) may serve as the final answer for single-step tasks or as an intermediate result in multi-step settings, where the process continues until the overall task is completed.

This unified formulation expands the action space of ReAct-style \citep{yao2023react} agents by introducing an explicit memory reasoning mechanism. 
Through this extension, memory becomes an adaptive component that interacts with reasoning in real time rather than remaining a passive context. Under this view, the entire decision loop can also be interpreted as a Markov process, where the state encapsulates the current input, memory state, and ongoing reasoning traces.
Such integration yields a lightweight yet powerful paradigm for continual adaptation, where the agent learns to reason about both the task and its own knowledge state. 
By coupling reflection with memory evolution, \alg establishes a new standard for adaptive, self-improving LLM agents.

\section{Experiments}
In this section, we evaluate leading LLMs on the \method benchmark under our unified test-time learning pipeline, focusing on five key research questions (RQs):
\begin{itemize}[itemsep=0pt,topsep=2pt,leftmargin=*]
    \item \textbf{RQ1:} How do LLM agents perform on \method across domains and task types, and does \textsc{ReMem} enhance their test-time learning ability?
    \item \textbf{RQ2:} What factors influence the effectiveness of memory in different tasks, and how does experience reuse improve task efficiency?
    \item \textbf{RQ3:} How does task sequence difficulty (e.g., easy vs.\ hard trajectories) affect memory adaptation and generalization?
    \item \textbf{RQ4:} How do varying feedback types impact learning dynamics and memory refinement?
\end{itemize}

\subsection{Experimental Setup}
\label{exp_setup}
\method evaluates memory mechanisms under realistic streaming multi-task conditions. In what follows, we describe the benchmark datasets, metrics, and the methods compared.

\subsubsection{Datasets}

\method is evaluated on a diverse suite of datasets spanning factual knowledge, reasoning, mathematics, programming, and goal-oriented interaction.  
For factual and reasoning ability, we include \textbf{MMLU-Pro}~\citep{zheng2024mmlu} and \textbf{GPQA-Diamond}~\citep{rein2023gpqa}, which test multi-disciplinary and graduate-level reasoning.  
For mathematical problem solving, we use \textbf{AIME-24} and \textbf{AIME-25}~\citep{aime2024}, containing math challenges requiring symbolic reasoning and exact-match evaluation.  
For tool-use and API grounding, we include \textbf{ToolBench}~\citep{patil2023gorilla}.  
For multi-turn and goal-oriented interaction, we adopt \textbf{AlfWorld}~\citep{shridhar2021alfworld}, \textbf{BabyAI}~\citep{chevalier2018babyai}, \textbf{ScienceWorld}~\citep{wang2022scienceworld}, and \textbf{PDDL} tasks~\citep{pddlbench}.  
All methods are evaluated under the same \emph{search–predict–evolve} loop
\[
(x_t, M_t) \xrightarrow{\text{search}} R_t \xrightarrow{\text{synthesis}} \hat{y}_t \xrightarrow{\text{evolve}} M_{t+1},
\]
with preferred prompting templates, and configurations. Feedback $f_t$ is considered the correctness signal.

\subsubsection{Methods}

We benchmark a broad range of agents and memory architectures instantiated on two strong \textbf{LLM backbones}: the Gemini-2.5 series~\citep{comanici2025gemini} (\textsc{Flash}, \textsc{Flash-Lite}, and \textsc{Pro}) and the Claude family~\citep{anthropic_claude4_2025} (\textsc{3.5-Haiku} and \textsc{3.7-Sonnet}).
The evaluated methods are grouped into four categories:  
(1) \textbf{Agent pipelines without procedural memory}, including ReAct~\citep{yao2023react} and Amem \citep{xu2025mem}, which rely on context instead of learned procedures.;  
(2) \textbf{Adaptive agentic memory methods}, such as SelfRAG~\citep{asai2024self}, MemOS~\citep{Li2025MemOSAO}, Mem0~\citep{mem0}, and LangMem~\citep{langchain}, which support dynamic retrieval and continual updates;  
(3) \textbf{Memory-based agents for procedural knowledge}, including Dynamic Cheatsheet (DC)~\citep{suzgun2025dynamic} with two variants Cumulative (Cu) and Synthesis (RS), and Agent Workflow Memory (AWM)~\citep{Wang2024AgentWM}, which emphasize reusable workflows and task strategies; and  
(4) \textbf{Proposed evolving-memory framework}, comprising \textbf{ExpRecent}, \textbf{ExpRAG}, and \textbf{ReMem}, which unify reasoning, action, and memory refinement in a self-evolving loop.  
All methods are evaluated under a unified \emph{search–predict–evolve} protocol to isolate the effects of memory design. Implementation and prompting details are provided in Appendix~\ref{appendix:setup}. We exclude systems such as MemoryGpt \citep{Zhong2023MemoryBankEL} and MemoryBank \citep{Zhong2023MemoryBankEL} that target factual recall only, as well as methods incompatible with embodied environments (e.g., MemOS and LangMem) from multi-turn evaluations.

\begin{table*}[t!]
\centering
\begin{minipage}[t]{0.5\textwidth}
\centering
\scriptsize
\setlength{\tabcolsep}{3.5pt}
\renewcommand{\arraystretch}{1.1}
\scalebox{0.6}{
\begin{tabular}{l l*{8}{c}}
\toprule
\multirow{3}{*}{\textbf{\shortstack{LLM\\Backbone}}} & \multirow{3}{*}{\textbf{Method}} & \multicolumn{6}{c}{\textbf{Exact Match $\uparrow$}} & \multicolumn{1}{c}{\textbf{API / Acc. $\uparrow$}}\\
\cmidrule(lr){3-8}\cmidrule(lr){9-9}
 & & \multirow{2}{*}{AIME24} & \multirow{2}{*}{AIME25} & \multirow{2}{*}{GPQA} & \multicolumn{3}{c}{MMLU-Pro} & \multirow{2}{*}{ToolBen.} & \multirow{2}{*}{Avg. $\uparrow$} \\
\cmidrule(lr){6-8} 
 & & & & & Eco. & Eng. & Philo. & & \\
\midrule

\multirow{15}{*}{\shortstack{Claude\\3.7\\Sonnet}}
 & Baseline            & 0.17 & 0.13 & 0.55 & 0.84 & 0.63 & 0.78 & 0.76/0.62 & 0.54 \\
 & History       & 0.13 & \textbf{0.23} & 0.56 & 0.85 & 0.64 & 0.78 & 0.76/0.61 & 0.55 \\
\cmidrule(lr){2-10}
 & ReAct                     & 0.17 & 0.10 & 0.57 & 0.84 & 0.63 & 0.76 & 0.76/0.61 & 0.54 \\
 & Amem                      & \textbf{0.27} & 0.17 & 0.54 & 0.83 & 0.63 & 0.79 & 0.77/0.63 & 0.56 \\
\cmidrule(lr){2-10}
 & SelfRAG                   & 0.20 & 0.10 & 0.58 & 0.84 & 0.65 & 0.77 & 0.77/0.63 & 0.55 \\
 & MemOS                     & 0.17 & 0.20 & 0.55 & 0.84 & 0.64 & 0.76 & 0.76/0.62 & 0.55 \\
 & Mem0                      & 0.20 & 0.13 & 0.58 & 0.84 & 0.62 & 0.77 & 0.76/0.61 & 0.55 \\
 & LangMem                   & 0.10 & 0.13 & 0.53 & 0.77 & 0.56 & 0.66 & 0.77/0.63 & 0.49 \\
\cmidrule(lr){2-10}
 & DC-Cu                     & 0.17 & \textbf{0.23} & 0.57 & 0.79 & 0.52 & 0.65 & 0.77/0.62 & 0.52 \\
 & DC-RS                     & 0.20 & 0.20 & 0.62 & 0.79 & 0.52 & 0.60 & 0.77/0.62 & 0.52 \\
 & AWM                       & 0.03 & 0.03 & 0.53 & 0.80 & 0.56 & 0.72 & 0.76/0.62 & 0.48 \\
\cmidrule(lr){2-10}
 & ExpRecent                 & 0.13 & 0.20 & 0.61 & \textbf{0.86} & 0.63 & 0.78 & 0.82/0.66 & 0.56 \\
 & ExpRAG                    & 0.17 & 0.17 & \textbf{0.70} & 0.85 & \textbf{0.67} & \textbf{0.80} & \textbf{0.88/0.72} & \textbf{0.59} \\
 & ReMem                     & 0.13 & 0.13 & 0.67 & \textbf{0.86} & 0.65 & 0.80 & 0.87/0.71 & 0.58 \\
\midrule

\multirow{15}{*}{\shortstack{Gemini\\2.5\\Flash}}
 & Baseline           & 0.47 & 0.47 & 0.48 & 0.83 & \textbf{0.46} & 0.75 & 0.71/0.61 & 0.59 \\
 & History       & 0.60 & 0.47 & 0.43 & 0.84 & 0.42 & 0.78 & 0.62/0.54 & 0.58 \\
\cmidrule(lr){2-10}
 & ReAct                    & 0.30 & 0.27 & 0.05 & 0.64 & 0.16 & 0.54 & 0.64/0.57 & 0.37 \\
 & Amem                     & \textbf{0.70} & \textbf{0.57} & 0.52 & 0.83 & 0.42 & 0.72 & 0.72/0.60 & 0.63 \\
\cmidrule(lr){2-10}
 & SelfRAG                  & 0.50 & 0.47 & 0.46 & 0.83 & 0.45 & 0.75 & 0.72/0.61 & 0.59 \\
 & MemOS                    & 0.47 & 0.47 & 0.50 & 0.82 & \textbf{0.46} & 0.75 & 0.71/0.61 & 0.59 \\
 & Mem0                     & 0.50 & 0.47 & 0.45 & 0.83 & \textbf{0.46} & 0.74 & 0.71/0.61 & 0.59 \\
 & LangMem                  & 0.43 & 0.50 & \textbf{0.53} & 0.79 & 0.39 & 0.71 & 0.68/0.57 & 0.57 \\
\cmidrule(lr){2-10}
 & DC-Cu                    & 0.60 & 0.40 & 0.48 & 0.79 & 0.44 & 0.69 & 0.70/0.59 & 0.58 \\
 & DC-RS                    & 0.53 & 0.37 & 0.48 & 0.80 & 0.42 & 0.69 & 0.68/0.57 & 0.56 \\
 & AWM                      & 0.50 & 0.37 & 0.49 & 0.79 & 0.43 & 0.72 & 0.71/0.59 & 0.56 \\
\cmidrule(lr){2-10}
 & ExpRecent                & 0.47 & 0.47 & 0.42 & 0.83 & 0.39 & 0.75 & 0.78/0.66 & 0.58 \\
 & ExpRAG                   & 0.43 & 0.47 & 0.42 & 0.83 & 0.43 & 0.78 & \textbf{0.87/0.73} & 0.60 \\
 & ReMem                    & 0.60 & 0.53 & 0.51 & \textbf{0.85} & \textbf{0.46} & \textbf{0.79} & 0.85/0.71 & \textbf{0.65} \\
\bottomrule
\end{tabular}}
\subcaption{Single-turn reasoning and QA results.}
\label{tab:single_part}
\end{minipage}
\hfill
\begin{minipage}[t]{0.48\textwidth}
\centering
\scriptsize
\setlength{\tabcolsep}{4pt}
\renewcommand{\arraystretch}{1.2}
\scalebox{0.65}{
\begin{tabular}{l l cc cc cc cc cc}
\toprule
\multirow{2}{*}{\textbf{\shortstack{LLM\\Backbone}}} & \multirow{2}{*}{\textbf{Method}} 
& \multicolumn{2}{c}{\textbf{AlfWorld}} 
& \multicolumn{2}{c}{\textbf{BabyAI}} 
& \multicolumn{2}{c}{\textbf{PDDL}} 
& \multicolumn{2}{c}{\textbf{ScienceWorld}} 
& \multicolumn{2}{c}{\textbf{Avg.}} \\
\cmidrule(lr){3-4}\cmidrule(lr){5-6}\cmidrule(lr){7-8}\cmidrule(lr){9-10}\cmidrule(lr){11-12}
 &  & S & P & S & P & S & P & S & P & S & P \\
\midrule
\multirow{15}{*}{\shortstack{Gemini \\ 2.5 \\ Flash}}
 & Baseline & 0.12 & 0.34 & \textbf{0.61} & \textbf{0.71} & 0.12 & 0.20 & 0.24 & 0.59 & 0.27 & 0.46 \\
 & History & 0.28 & 0.60 & 0.52 & 0.64 & 0.08 & 0.15 & 0.31 & 0.71 & 0.30 & 0.53 \\
\cmidrule(lr){2-12}
 & ReAct & 0.24 & 0.56 & 0.48 & 0.63 & \textbf{0.22} & \textbf{0.33} & 0.34 & 0.71 & 0.32 & 0.56 \\
 & Amem & 0.25 & 0.59 & 0.53 & 0.64 & 0.10 & 0.16 & 0.36 & 0.74 & 0.31 & 0.53 \\
\cmidrule(lr){2-12}
 & SelfRAG & 0.25 & 0.59 & 0.52 & 0.65 & 0.08 & 0.16 & 0.34 & 0.74 & 0.30 & 0.54 \\
 & Mem0 & 0.27 & 0.61 & 0.54 & 0.66 & 0.10 & 0.19 & 0.32 & 0.70 & 0.31 & 0.54 \\
\cmidrule(lr){2-12}
 & DC-Cu & 0.25 & 0.59 & 0.53 & 0.64 & 0.08 & 0.17 & 0.29 & 0.71 & 0.29 & 0.53 \\
 & DC-RS & 0.27 & 0.60 & 0.53 & 0.66 & 0.07 & 0.15 & 0.33 & 0.73 & 0.30 & 0.54 \\
 & AWM & 0.26 & 0.59 & 0.52 & 0.64 & 0.08 & 0.16 & 0.33 & 0.73 & 0.30 & 0.53 \\
\cmidrule(lr){2-12}
 & ExpRecent & 0.37 & 0.65 & 0.53 & 0.64 & 0.13 & 0.22 & 0.53 & \textbf{0.83} & 0.39 & 0.59 \\
 & ExpRAG & 0.59 & 0.79 & 0.56 & 0.65 & 0.17 & 0.27 & 0.53 & 0.81 & 0.46 & 0.63 \\
 & ReMem & \textbf{0.66} & \textbf{0.81} & 0.53 & 0.61 & \textbf{0.22} & \textbf{0.33} & \textbf{0.58} & 0.81 & \textbf{0.50} & \textbf{0.64} \\
\midrule
\multirow{15}{*}{\shortstack{Claude\\3.7\\Sonnet}}
 & Baseline & 0.18 & 0.49 & 0.51 & 0.66 & 0.17 & 0.39 & 0.10 & 0.53 & 0.24 & 0.52 \\
 & History & 0.50 & 0.73 & 0.48 & 0.66 & 0.65 & 0.85 & 0.32 & 0.74 & 0.49 & 0.74 \\
\cmidrule(lr){2-12}
 & ReAct & 0.51 & 0.75 & 0.57 & 0.72 & 0.75 & 0.91 & 0.44 & 0.77 & 0.57 & 0.79 \\
 & Amem & 0.48 & 0.73 & 0.46 & 0.64 & 0.62 & 0.84 & 0.33 & 0.73 & 0.47 & 0.73 \\
\cmidrule(lr){2-12}
 & SelfRAG & 0.52 & 0.75 & 0.46 & 0.64 & 0.65 & 0.84 & 0.31 & 0.74 & 0.49 & 0.74 \\
 & Mem0 & 0.51 & 0.74 & 0.48 & 0.66 & 0.65 & 0.84 & 0.37 & 0.76 & 0.50 & 0.75 \\
\cmidrule(lr){2-12}
 & DC-Cu & 0.50 & 0.74 & 0.50 & 0.67 & 0.62 & 0.84 & 0.33 & 0.75 & 0.49 & 0.75 \\
 & DC-RS & 0.50 & 0.74 & 0.52 & 0.68 & 0.62 & 0.84 & 0.34 & 0.74 & 0.50 & 0.75 \\
 & AWM & 0.49 & 0.73 & 0.53 & 0.68 & 0.60 & 0.82 & 0.34 & 0.74 & 0.49 & 0.74 \\
\cmidrule(lr){2-12}
 & ExpRecent & 0.66 & 0.83 & 0.63 & 0.73 & 0.53 & 0.76 & 0.49 & 0.82 & 0.58 & 0.79 \\
 & ExpRAG & 0.74 & 0.89 & 0.62 & 0.72 & 0.72 & 0.89 & 0.46 & 0.76 & 0.63 & 0.82 \\
 & ReMem & \textbf{0.92} & \textbf{0.96} & \textbf{0.73} & \textbf{0.83} & \textbf{0.83} & \textbf{0.95} & \textbf{0.62} & \textbf{0.89} & \textbf{0.78} & \textbf{0.91} \\
\bottomrule
\end{tabular}}
\subcaption{Multi-turn embodied reasoning results.}
\label{tab:multi_part}
\end{minipage}
\caption{Cross-benchmark results of diverse memory architectures across single-turn and multi-turn tasks. \textbf{Left:} single-turn reasoning and question answering results. \textbf{Right:} multi-turn embodied reasoning results.}
\label{tab:combined_results}
\end{table*}

\subsection{Experimental Results}

Below are the conducted experiments to answer the proposed research questions.

% \input{tables/single_step_table_full}

% \input{tables/multi_step_table_full}

% \begin{table*}[t]
% \centering
% \small
% \renewcommand{\arraystretch}{1.15}
% \setlength{\tabcolsep}{8pt}
% \begin{tabular}{lccccc}
% \hline
% \textbf{Method} & \textbf{Alfworld} & \textbf{BabyAI} & \textbf{Jericho} & \textbf{PDDL} & \textbf{ScienceWorld} \\
% \hline
% Baseline & 0.12 & 0.61 & 0.05 & 0.12 & 0.24 \\
% Full History & 0.28 & 0.52 & 0.30 & 0.08 & 0.31 \\
% \hline
% \multicolumn{6}{l}{\textit{Agentic Solving Pipeline}} \\
% ReAct & 0.24 & 0.48 & 0.30 & 0.22 & 0.34 \\
% Amem  & 0.26 & 0.53 & 0.40 & 0.10 & 0.36 \\
% \hline
% \multicolumn{6}{l}{\textit{Adaptive agentic memory methods}} \\
% SelfRAG & 0.26 & 0.52 & 0.40 & 0.08 & 0.34 \\
% Mem0    & 0.27 & 0.54 & 0.30 & 0.10 & 0.32 \\
% \hline
% \multicolumn{6}{l}{\textit{Memory-based agents for procedural memory}} \\
% DC-Cu (Dynamic Cheatsheet) & 0.25 & 0.53 & 0.30 & 0.08 & 0.29 \\
% DC-RS (Dynamic Cheatsheet) & 0.27 & 0.53 & 0.30 & 0.07 & 0.33 \\
% AWM                        & 0.26 & 0.52 & 0.25 & 0.08 & 0.33 \\
% \hline
% \multicolumn{6}{l}{\textit{Proposed agentic memory}} \\
% ExpRecent               & 0.37 & 0.53 & 0.30 & 0.13 & 0.53 \\
% ExpRAG & 0.66 & 0.56 & 0.40 & 0.12 & 0.56 \\
% ReMem &  &  &  &  & \\
% \hline
% \end{tabular}
% \caption{Results across datasets (higher is better). Ordering follows the figure. Name mapping: \emph{vanilla} $\rightarrow$ Full-history; \emph{vanilla\_no} $\rightarrow$ Baseline; \emph{fewshot\_rag} $\rightarrow$ SelfStream.}
% \label{tab:agentic-memory-results}
% \end{table*}

\subsubsection{Analysis of Results (RQ1)}

Tables~\ref{tab:single_part} and~\ref{tab:multi_part} summarize the results across single-turn and multi-turn settings. 
Overall, \method demonstrates that self-evolving memory architectures provide consistent improvements. In single-turn reasoning and QA benchmarks (AIME-24/25, GPQA, MMLU-Pro, ToolBench), evolving-memory methods show consistent improvements, with \textsc{ReMem} achieving 0.65 average exact match and 0.85/0.71 API accuracy under Gemini-2.5~Flash. 
Adaptive retrieval methods enhance factual grounding, yet only evolving systems maintain consistent gains through iterative refinement. 
Agents with procedural knowledge perform well on structured domains such as AIME but lag in scientific reasoning and tool use, showing limited flexibility. \textsc{ExpRAG} serves as a simple yet highly effective baseline, outperforming several more complex designs.
While improvements in single-turn settings are moderate, the overall trend remains consistent across datasets and model families.

In multi-turn reasoning environments (AlfWorld, BabyAI, PDDL, ScienceWorld), \textsc{ReMem} and \textsc{ExpRAG} achieve strong and stable performance on both Gemini-2.5 and Claude backbones, reaching 0.92/0.96 on BabyAI and 0.95/0.62 on ScienceWorld.
These results indicate that continual reflection and refinement substantially improve procedural knowledge accumulation.
Performance gains are notably larger in multi-turn settings, underscoring that continual adaptation becomes increasingly valuable as task horizons lengthen. While many baselines enhance retrieval grounding, they struggle to reuse long-horizon experiences and often falter in open-ended environments. Notably, lightweight variants such as \textsc{ExpRecent} and \textsc{ExpRAG} still perform competitively despite their simplicity, suggesting that explicit task-level utilization during test-time evolution is both promising and underexplored.

% Across all experiments, evolving-memory methods demonstrate consistent gains on both Gemini and Claude backbones. 
% Smaller models benefit particularly from self-evolving memory, suggesting that test-time refinement is a practical path to enhancing the capability of lighter LLMs.
% Together, these findings establish task-level memory utilization and continual reorganization as valuable directions for future research, providing a standardized reference point for developing and evaluating evolving-memory agents. Additional results across more LLM families are presented in Appendix \ref{app:exp}.

Across all experiments, evolving-memory methods consistently improve performance on various backbones and tasks. 
These results highlight task-level memory utilization and continual reorganization as promising directions for evolving-memory agents. 
Additional results across more LLM families are reported in Appendix~\ref{app:exp}. Smaller models benefit most, indicating that test-time refinement is also an effective way to enhance lightweight LLMs.

\subsubsection{Analysis of Memory Improvement (RQ2)}

\begin{figure}[t]
    \centering
    \includegraphics[width=0.83\linewidth]{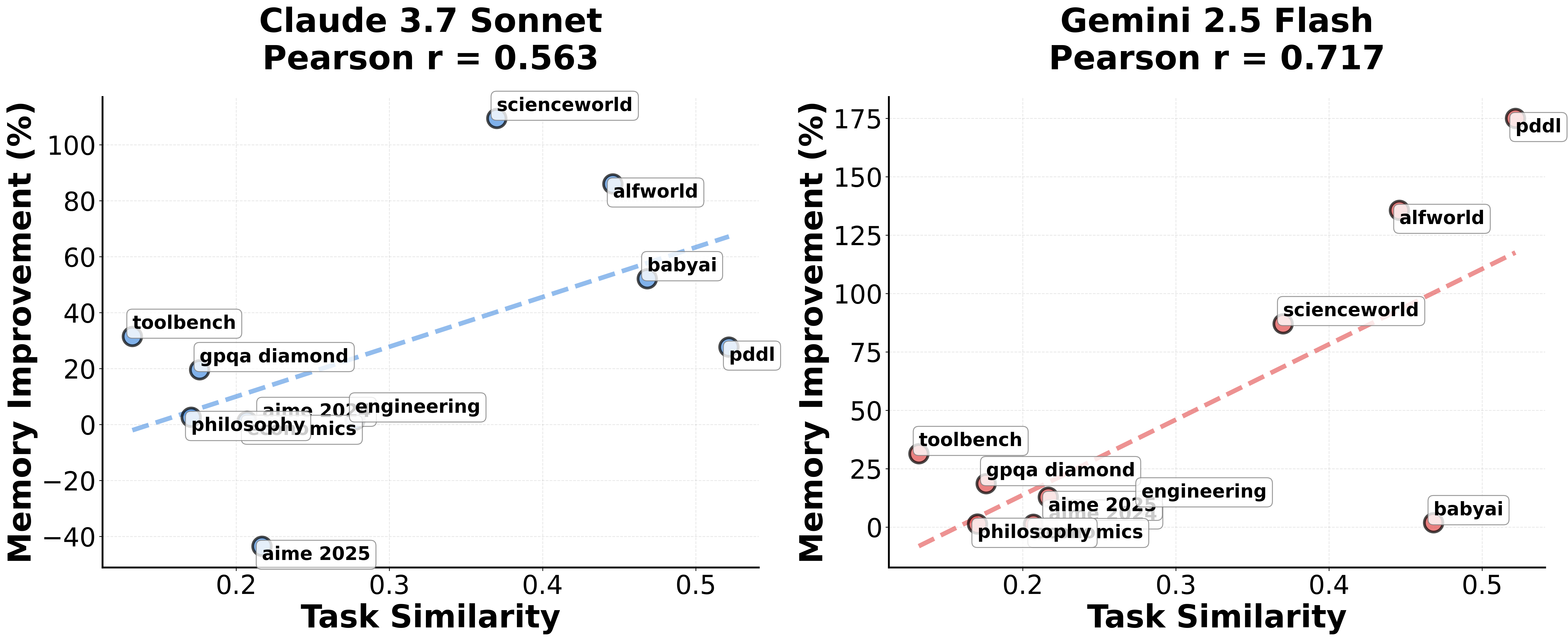}
    \caption{ReMem performance gain over history baseline versus within-dataset task similarity.}
    \label{fig:correlate}
\end{figure}

Figure~\ref{fig:correlate} shows that \textsc{ReMem}'s improvement strongly correlates with within-dataset task similarity (Pearson $r=0.717$ on Gemini~2.5~Flash and $r=0.563$ on Claude~3.7~Sonnet).
Task similarity is measured by computing the average cosine distance between each task embedding and its dataset cluster center, where embeddings are obtained from the retriever encoder.
A smaller average distance indicates higher intra-dataset coherence and thus stronger structural similarity.
Tasks with higher embedding cluster ratios, such as PDDL and AlfWorld, yield larger gains, suggesting that recurring task structures facilitate memory reuse and generalization.
In contrast, more diverse or low-similarity datasets like AIME-25 or GPQA show smaller gains, reflecting limited transferable experiences.
These findings highlight the importance of embedding organization and semantic overlap in driving effective memory evolution. Further analysis of memory pruning rates can be found in Appendix~\ref{app:memory_prune}.

% Figure~\ref{fig:step} compares step efficiency across four environments. 
% Evolving-memory methods consistently require fewer steps to reach completion, with \textsc{ReMem} achieving the strongest and most stable reductions (e.g., from 22.6 to 11.5 steps on AlfWorld). 
% The lightweight \textsc{ExpRAG} and \textsc{ExpRecent} also perform competitively, showing that simple task-level evolution can greatly improve efficiency without extra complexity. 
% Overall, continual refinement not only boosts accuracy but also makes reasoning more focused and efficient.

\begin{figure}[t!]
    \centering
    \includegraphics[width=0.7\linewidth]{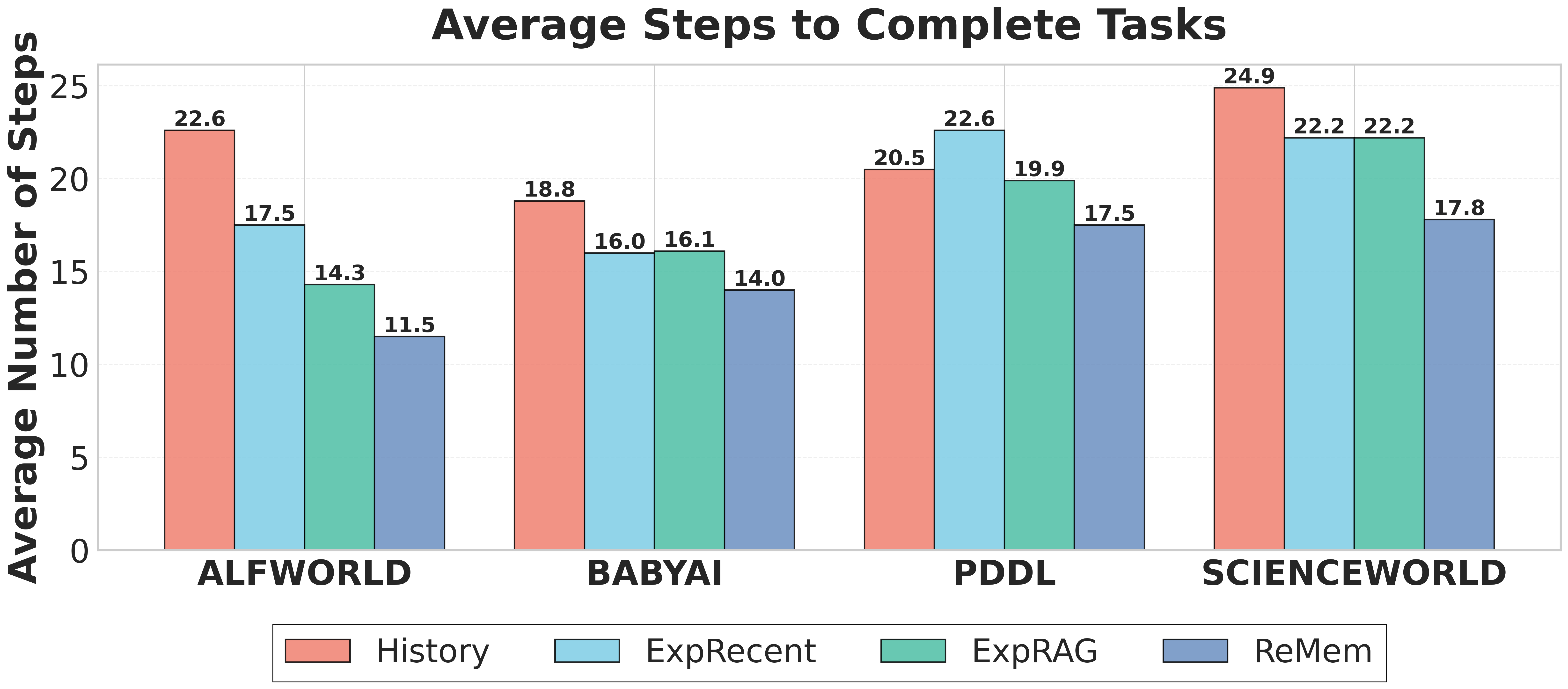}
    \caption{Average steps to complete tasks across four benchmarks. We compare four methods: History, ExpRecent, ExpRAG, and ReMem. Lower is better.}
    \label{fig:step}
\end{figure}

Figure~\ref{fig:step} compares step efficiency across four environments. 
Evolving-memory methods consistently require fewer steps, with \textsc{ReMem} achieving the largest and most stable reductions (e.g., from 22.6 to 11.5 steps on AlfWorld). 
Lightweight methods such as \textsc{ExpRAG} and \textsc{ExpRecent} also perform competitively, indicating that simple task-level evolution substantially improves efficiency. 
Overall, continual refinement leads to more focused and efficient reasoning.

\begin{table}[t!]
\centering
\small
\setlength{\tabcolsep}{4pt}
\renewcommand{\arraystretch}{1.2}
\scalebox{0.8}{
\begin{tabular}{llccccccc}
\toprule
\multirow{2}{*}{\textbf{Direction}} & \multirow{2}{*}{\textbf{Method}} 
& \multicolumn{2}{c}{\textbf{AlfWorld}} 
& \multicolumn{2}{c}{\textbf{ScienceWorld}} 
& \multicolumn{2}{c}{\textbf{Avg.}} \\
\cmidrule(lr){3-4} \cmidrule(lr){5-6} \cmidrule(lr){7-8}
 &  & S & P & S & P & S & P \\
\midrule
\multicolumn{2}{c}{Base} & 0.50 & 0.73 & 0.32 & 0.74 & 0.41 & 0.74 \\
\midrule
\multirow{3}{*}{Easy→Hard} 
 & ExpRecent & 0.66 & 0.82 & 0.48 & 0.83 & 0.57 & 0.83 \\
 & ExpRAG & 0.77 & 0.87 & 0.37 & 0.71 & 0.57 & 0.79 \\
 & ReMem & \textbf{0.91} & \textbf{0.96} & \textbf{0.63} & \textbf{0.88} & \textbf{0.77} & \textbf{0.92} \\
\midrule
\multirow{3}{*}{Hard→Easy} 
 & ExpRecent & 0.72 & 0.85 & 0.47 & 0.80 & 0.60 & 0.83 \\

 & ExpRAG & 0.87 & 0.92 & 0.51 & 0.81 & 0.69 & 0.87 \\
 & ReMem & \textbf{0.94} & \textbf{0.97} & \textbf{0.68} & \textbf{0.90} & \textbf{0.81} & \textbf{0.94} \\
\bottomrule
\end{tabular}}
\caption{Comparison of memory-based agents under different sequence difficulty directions. Easy→Hard and Hard→Easy indicate task order transitions.}
\label{tab:seq_difficulty}
\end{table}

\subsubsection{Task Sequence: Easy v.s. Hard (RQ3)}

Table~\ref{tab:seq_difficulty} examines how memory-based agents adapt to changes in task difficulty. 
Baseline methods exhibit substantial variation across task sequences, indicating limited robustness under distribution shifts. 
In contrast, evolving-memory agents, particularly \textsc{ReMem}, maintain strong and consistent performance in both directions, reaching up to 0.94/0.97 success and progress in the Hard$\rightarrow$Easy setting. 
This asymmetry across sequences also suggests that successful experiences from harder tasks are more transferable, contributing to the improved performance of \textsc{ReMem}. 
Overall, these results show that continual reflection helps retain transferable knowledge as task complexity varies, and highlight the importance of task sequence design for fair evaluation and effective learning.

\subsubsection{Analysis of Feedback (RQ4)}
% Table~\ref{tab:multi_norm} evaluates how agents perform when both successful and failed task experiences are stored in memory.
% Baseline methods experience a clear performance drop when exposed to unfiltered failures, indicating that naive memory accumulation introduces noise and disrupts subsequent retrieval.
% In contrast, evolving-memory approaches, particularly \textsc{ReMem}, remain robust by actively refining stored experiences, achieving the highest overall success and progress rates under both Claude and Gemini backbones.
% These results demonstrate that selective utilization, which involves learning from successes while appropriately leveraging information from failures, is crucial for stable test-time adaptation.
% They further highlight that memory refinement plays a central role in handling imperfect experiences and suggest that future work should explore failure-aware strategies for memory evolution.

\begin{table}[t!]
\centering
\small
\setlength{\tabcolsep}{6pt}
\renewcommand{\arraystretch}{1.15}
\scalebox{0.65}{
\begin{tabular}{l l cc cc cc}
\toprule
\textbf{Model} & \textbf{Method} & \multicolumn{2}{c}{\textbf{AlfWorld}} & \multicolumn{2}{c}{\textbf{ScienceWorld}} & \multicolumn{2}{c}{\textbf{Avg.}} \\
\cmidrule(lr){3-4}\cmidrule(lr){5-6}\cmidrule(lr){7-8}
 &  & S & P & S & P & S & P \\
\midrule
\multirow{9}{*}{Claude 3.7 Sonnet}
 & Amem        & 0.49 & 0.73 & 0.31 & 0.74 & 0.40 & 0.74 \\
\cmidrule(lr){2-8}
 & SelfRAG     & 0.47 & 0.73 & 0.34 & 0.73 & 0.41 & 0.73 \\
 & Mem0        & 0.49 & 0.73 & 0.36 & 0.74 & 0.43 & 0.74 \\
\cmidrule(lr){2-8}
 & DC-Cu       & 0.52 & 0.75 & 0.34 & 0.73 & 0.43 & 0.74 \\
 & DC-RS       & 0.51 & 0.74 & 0.38 & 0.74 & 0.45 & 0.74 \\
 & AWM         & 0.55 & 0.76 & 0.32 & 0.72 & 0.44 & 0.74 \\
\cmidrule(lr){2-8}
 & ExpRecent   & 0.62 & 0.80 & 0.34 & 0.74 & 0.48 & 0.77 \\
 & ExpRAG      & 0.76 & 0.90 & 0.27 & 0.63 & 0.52 & 0.77 \\
 & ReMem       & \textbf{0.92} & \textbf{0.96} & \textbf{0.69} & \textbf{0.91} & \textbf{0.81} & \textbf{0.94} \\
\midrule
\multirow{9}{*}{Gemini 2.5 Flash}
 & Amem        & 0.22 & 0.57 & 0.39 & 0.75 & 0.31 & 0.66 \\
\cmidrule(lr){2-8}
 & SelfRAG     & 0.25 & 0.58 & 0.36 & 0.71 & 0.31 & 0.65 \\
 & Mem0        & 0.25 & 0.59 & 0.34 & 0.71 & 0.30 & 0.65 \\
\cmidrule(lr){2-8}
 & DC-Cu       & 0.20 & 0.56 & 0.36 & 0.72 & 0.28 & 0.64 \\
 & DC-RS       & 0.21 & 0.57 & 0.36 & 0.71 & 0.29 & 0.64 \\
 & AWM         & 0.19 & 0.56 & 0.36 & 0.74 & 0.28 & 0.65 \\
\cmidrule(lr){2-8}
 & ExpRecent   & 0.22 & 0.57 & \textbf{0.59} & \textbf{0.86} & 0.41 & 0.72 \\
 & ExpRAG      & 0.25 & 0.60 & 0.51 & 0.78 & 0.38 & 0.69 \\
 & ReMem       & \textbf{0.57} & \textbf{0.76} & 0.50 & 0.75 & \textbf{0.54} & \textbf{0.76} \\
\bottomrule
\end{tabular}}
\caption{Results with both successful and failed task experiences on AlfWorld and ScienceWorld. Bold numbers denote the best results per metric.}
\label{tab:multi_norm}
\end{table}

Table~\ref{tab:multi_norm} evaluates agent performance when both successful and failed experiences are stored in memory. 
Baseline methods suffer notable degradation under unfiltered failures, indicating that naive memory accumulation introduces noise and hinders retrieval. 
In contrast, evolving-memory approaches, particularly \textsc{ReMem}, remain robust by actively refining stored experiences, achieving the best overall success and progress rates across both Claude and Gemini backbones. 
These results highlight the importance of selective utilization and memory refinement for stable test-time adaptation, and motivate future work on failure-aware memory evolution.

% \begin{figure}
%     \centering
%     \includegraphics[width=0.9\linewidth]{figures/rolling.pdf}
%     \caption{Rolling accuracy of the memory methods.}
%     \label{fig:placeholder}
% \end{figure}

\section{Conclusion}
Self-evolving memory is a fundamental yet underexplored aspect of LLM capability. While prior work centers on static conversational recall, it overlooks how models accumulate and reuse experience across evolving task streams. \method fills this gap by transforming static datasets into streaming trajectories, systematically evaluating how LLMs retrieve, adapt, and refine memory through interaction. Our results show that memory can substantially enhance performance but remains fragile in stability and procedural reuse. To foster progress, we introduce ExpRAG for experience retrieval and \alg for interleaving reasoning, action, and memory updates. We hope \method serves as a unified platform for building LLMs with reliable and continually improving memory.

\bibliography{main}

\newpage
\clearpage
\appendix
\textbf{\Large Appendix}

\addtocontents{toc}{\protect\setcounter{tocdepth}{2}}

\tableofcontents
\clearpage
\section{Experimental Details}
\label{appendix:setup}

\method evaluates memory mechanisms under realistic streaming multi-task conditions. In what follows, we describe the benchmark datasets, metrics, configurations and the methods compared in details.

\subsection{Datasets}
\label{app:dataset}
We evaluate our approach on a diverse suite of benchmarks that span factual knowledge, reasoning, mathematics, programming, and goal-oriented interaction.

We first introduce a suite of single-turn datasets designed to evaluate diverse reasoning abilities. \textbf{MMLU-Pro}~\citep{zheng2024mmlu} extends the original MMLU benchmark with stronger robustness and challenge by filtering data leakage, reducing ambiguity, and introducing more difficult questions across domains such as engineering, philosophy, and economics, making it a more reliable testbed for assessing multi-disciplinary reasoning. \textbf{GPQA-Diamond}~\citep{rein2023gpqa} is a graduate-level benchmark featuring expert-written, ``Google-proof'' questions in physics and related sciences, with its Diamond split being the most challenging and requiring rigorous multi-step reasoning. \textbf{AIME-24} and \textbf{AIME-25}~\citep{aime2024,aime2025} consist of Olympiad-style mathematics problems from the 2024 and 2025 American Invitational Mathematics Examinations, testing symbolic manipulation and problem-solving under strict exact-match criteria. Finally, \textbf{ToolBench}~\citep{patil2023gorilla} assesses a model’s ability to identify and configure external APIs, reflecting practical tool-use capabilities.

We then evaluate on a suite of multi-turn, goal-oriented benchmarks designed to evaluate memory in embodied and interactive environments. It includes several representative domains:
\textbf{AlfWorld}~\citep{shridhar2021alfworld} for household instruction following, 
\textbf{BabyAI}~\citep{chevalier2018babyai} for grounded navigation and compositional reasoning,
\textbf{ScienceWorld}~\citep{wang2022scienceworld} for open-ended scientific experimentation,
and \textbf{PDDL} tasks~\citep{pddlbench} for symbolic planning. 
Together, these environments emphasize long-horizon reasoning, sequential decision-making, and the use of accumulated experience to achieve complex goals.

Together, these datasets form a comprehensive benchmark suite that evaluates factual recall, domain expertise, mathematical reasoning, and procedural memory in interactive settings. This diversity enables a unified evaluation of both static and evolving capabilities, reflecting how LLMs learn, act, and adapt across academic and real-world scenarios.

\subsection{Configuration}
\label{app:config}
For efficient retrieval and fair comparison across methods, we utilize the BAAI/bge-base-en-v1.5~\citep{chen2024m3} encoder as the retriever to index both queries and memory items.
During inference, the current question is encoded as a query and compared with all stored memory embeddings, retrieving the top-$k$ most relevant items (default $k=4$) for contextual augmentation.
This setting ensures a consistent retrieval budget across all methods.
For efficiency, retrieved texts and task inputs are truncated to fit within the same prompt length constraint used by the generation models.

While all baselines adopt the same retrieval configuration, certain methods (e.g., \textsc{Self-RAG}, \textsc{ReMem}) introduce additional reasoning modules that determine \emph{whether to retrieve} and \emph{what to retrieve} at each step. These adaptive behaviors operate on top of the same retrieval pool to ensure comparability.

Across all experiments, we maintain a unified task sequence ordering within each dataset, ensuring consistent memory evolution dynamics for all models. Unless otherwise specified, retrieval and generation operate within the same pipeline, and the retrieved items are appended to the prompt following the order of relevance, from most to least similar.

We benchmark a broad range of agents and memory architectures instantiated on two strong \textbf{LLM backbones}: the Gemini-2.5 series~\citep{comanici2025gemini} (\textsc{Flash}, \textsc{Flash-Lite}, and \textsc{Pro}) and the Claude family~\citep{anthropic_claude4_2025} (\textsc{3.5-Haiku} and \textsc{3.7-Sonnet}).

\subsection{Evaluation}

\method evaluates both task performance and memory quality along four key dimensions: \begin{itemize}[itemsep=-0.2em, topsep=-0.3em, leftmargin=*] \item \textbf{Answer accuracy.} Evaluates whether the LLM produces correct outputs across tasks, reflecting its ability to incorporate past experiences into inference. \item \textbf{Success rate.} Measures whether the LLM agent successfully completes task goals, indicating its overall effectiveness in interactive or goal-oriented settings. \item \textbf{Step efficiency.} Tracks the number of steps required to complete a goal, assessing whether memory usage enables concise and scalable reasoning. \item \textbf{Sequence robustness.} Examines whether the LLM maintains consistent knowledge and performance across varying task orders, reflecting its ability to stably reuse prior experiences. \end{itemize}

\subsection{Methods}
\label{app:method}
We benchmark \method with a wide spectrum of agent and memory architectures to study how different designs impact \emph{test-time memory evolution}. All methods are instantiated on two strong \textbf{LLM backbones}: Gemini-2.5~\citep{comanici2025gemini} and Claude-3.5/3.7~\citep{anthropic_claude4_2025}. Our comparisons isolate the impact of memory architecture and update strategy. Differences in backbone capability are not the focus of the study. We group the evaluated approaches into four major families:

\paragraph{Agent Pipelines without Procedural Memory.}
\textbf{ReAct}~\citep{yao2023react} serves as a representative reasoning–action pipeline, where memory is limited to the immediate context. It generates interleaved reasoning traces and tool calls but does not explicitly store or evolve information.  
\textbf{Amem} \cite{xu2025mem} extends this pipeline with a lightweight agentic memory that caches recent observations and reflections. It provides a minimal form of experience reuse without dedicated search or update policies, forming a bridge between memory-free agents and adaptive memory systems.

\paragraph{Adaptive Agentic Memory Methods.}
This group focuses on adaptive retrieval and self-evolving memory. \textbf{SelfRAG}~\citep{asai2023self} integrates dynamic retrieval and reflection to adaptively ground reasoning in prior contexts.  
\textbf{MemOS}~\citep{Li2025MemOSAO}, \textbf{Mem0}~\citep{mem0}, and \textbf{LangMem}~\citep{langchain} implement structured, agent-level memory systems that support read, write, and update operations. Within our unified interface, retrieval corresponds to the \emph{search} stage and updates correspond to \emph{evolve}. These methods represent adaptive long-term agents capable of continual refinement.

\paragraph{Memory-Based Agents for Procedural Memory.}
\textbf{Dynamic Cheatsheet (DC)}~\citep{suzgun2025dynamic} and \textbf{Agent Workflow Memory (AWM)}~\citep{Wang2024AgentWM} emphasize the reuse of procedural knowledge, encoding “how-to” information rather than static facts.  
We evaluate two DC variants, \textbf{DC-RS} (retrieval-based) and \textbf{DC-Cu} (curated), to analyze how workflow induction and update mechanisms influence stability and transfer. These methods test the potential of procedural memory as reusable strategy repositories.

\paragraph{Proposed: Evolving Memory Framework.}
\textbf{ExpRecent} maintains condensed episodic traces of recent task trajectories, while our \textbf{ExpRAG} family integrates the principles of retrieval-augmented reasoning with explicit \emph{test-time evolution}. \textbf{ReMem} applies iterative reflection and synthesis to refine memory embeddings over time.  
Together, these methods instantiate \method's design philosophy, treating reasoning, acting, and memory refinement as interleaved processes that co-adapt during deployment, enabling continual self-improvement and more human-like adaptation.

\section{Experiments}

We provide more experiments in the following.
\subsection{Additional Experiments}
\label{app:exp}
We further validate our findings through extensive benchmarking across multiple model families (Gemini-2.5-Flash-Lite, Claude-3.5-Haiku) and diverse datasets, as shown in Tables~\ref{tab:multi_full} and~\ref{tab:single_full}.
The performance trends remain consistent across all settings.
On both multi-turn embodied reasoning tasks (AlfWorld, BabyAI, PDDL, ScienceWorld) and single-turn reasoning tasks (AIME-24/25, GPQA, MMLU-Pro, ToolBench), \alg consistently outperforms conventional baselines and adaptive retrieval methods across model backbones.
These results confirm that the advantages of evolving-memory architectures are model-agnostic, highlighting continual task-level reflection as a general mechanism for improving adaptability in problem-solving.

\begin{table*}[t!]
\centering
\scriptsize
\setlength{\tabcolsep}{4pt}
\renewcommand{\arraystretch}{1.2}
\scalebox{0.9}{
\begin{tabular}{l l cc cc cc cc cc}
\toprule
\multirow{2}{*}{\textbf{LLM Backbone}} & \multirow{2}{*}{\textbf{Method}} 
& \multicolumn{2}{c}{\textbf{AlfWorld}} 
& \multicolumn{2}{c}{\textbf{BabyAI}} 
& \multicolumn{2}{c}{\textbf{PDDL}} 
& \multicolumn{2}{c}{\textbf{ScienceWorld}} 
& \multicolumn{2}{c}{\textbf{Avg.}} \\
\cmidrule(lr){3-4}\cmidrule(lr){5-6}\cmidrule(lr){7-8}\cmidrule(lr){9-10}\cmidrule(lr){11-12}
 &  & S & P & S & P & S & P & S & P & S & P \\
\midrule
\multirow{15}{*}{Gemini 2.5 Flash}
 & Baseline & 0.12 & 0.34 & \textbf{0.61} & \textbf{0.71} & 0.12 & 0.20 & 0.24 & 0.59 & 0.27 & 0.46 \\
 & History & 0.28 & 0.60 & 0.52 & 0.64 & 0.08 & 0.15 & 0.31 & 0.71 & 0.30 & 0.53 \\
\cmidrule(lr){2-12}
 & ReAct & 0.24 & 0.56 & 0.48 & 0.63 & \textbf{0.22} & \textbf{0.33} & 0.34 & 0.71 & 0.32 & 0.56 \\
 & Amem & 0.25 & 0.59 & 0.53 & 0.64 & 0.10 & 0.16 & 0.36 & 0.74 & 0.31 & 0.53 \\
\cmidrule(lr){2-12}
 & SelfRAG & 0.25 & 0.59 & 0.52 & 0.65 & 0.08 & 0.16 & 0.34 & 0.74 & 0.30 & 0.54 \\
 & Mem0 & 0.27 & 0.61 & 0.54 & 0.66 & 0.10 & 0.19 & 0.32 & 0.70 & 0.31 & 0.54 \\
\cmidrule(lr){2-12}
 & DC-Cu & 0.25 & 0.59 & 0.53 & 0.64 & 0.08 & 0.17 & 0.29 & 0.71 & 0.29 & 0.53 \\
 & DC-RS & 0.27 & 0.60 & 0.53 & 0.66 & 0.07 & 0.15 & 0.33 & 0.73 & 0.30 & 0.54 \\
 & AWM & 0.26 & 0.59 & 0.52 & 0.64 & 0.08 & 0.16 & 0.33 & 0.73 & 0.30 & 0.53 \\
\cmidrule(lr){2-12}
 & ExpRecent & 0.37 & 0.65 & 0.53 & 0.64 & 0.13 & 0.22 & 0.53 & \textbf{0.83} & 0.39 & 0.59 \\
 & ExpRAG & 0.59 & 0.79 & 0.56 & 0.65 & 0.17 & 0.27 & 0.53 & 0.81 & 0.46 & 0.63 \\
 & \textbf{ReMem} & \textbf{0.66} & \textbf{0.81} & 0.53 & 0.61 & \textbf{0.22} & \textbf{0.33} & \textbf{0.58} & 0.81 & \textbf{0.50} & \textbf{0.64} \\
\midrule
\multirow{15}{*}{Gemini 2.5 Pro}
 & Baseline & 0.04 & 0.39 & 0.37 & 0.47 & 0.20 & 0.33 & 0.22 & 0.60 & 0.21 & 0.45 \\
 & History & 0.19 & 0.52 & 0.40 & 0.48 & 0.25 & 0.38 & 0.60 & 0.84 & 0.36 & 0.56 \\
\cmidrule(lr){2-12}
 & ReAct & 0.02 & 0.26 & 0.43 & 0.55 & 0.13 & 0.22 & 0.30 & 0.68 & 0.22 & 0.43 \\
 & Amem & 0.16 & 0.50 & 0.42 & 0.49 & 0.23 & 0.38 & 0.59 & 0.85 & 0.35 & 0.56 \\
\cmidrule(lr){2-12}
 & SelfRAG & 0.16 & 0.49 & 0.43 & 0.50 & 0.22 & 0.35 & 0.57 & 0.84 & 0.34 & 0.55 \\
 & Mem0 & 0.16 & 0.49 & 0.41 & 0.49 & 0.22 & 0.37 & 0.53 & 0.81 & 0.33 & 0.54 \\
\cmidrule(lr){2-12}
 & DC-Cu & 0.17 & 0.50 & 0.40 & 0.47 & 0.28 & 0.40 & 0.53 & 0.82 & 0.35 & 0.55 \\
 & DC-RS & 0.18 & 0.51 & 0.42 & 0.50 & 0.27 & 0.40 & 0.59 & 0.85 & 0.37 & 0.57 \\
 & AWM & 0.20 & 0.52 & 0.38 & 0.46 & 0.20 & 0.37 & 0.57 & 0.83 & 0.34 & 0.54 \\
\cmidrule(lr){2-12}
 & ExpRecent & 0.36 & 0.61 & 0.54 & \textbf{0.64} & \textbf{0.35} & \textbf{0.47} & \textbf{0.69} & \textbf{0.89} & 0.49 & 0.64 \\
 & ExpRAG & 0.38 & 0.64 & 0.46 & 0.53 & 0.28 & 0.43 & 0.61 & 0.84 & 0.43 & 0.61 \\
 & \textbf{ReMem} & \textbf{0.51} & \textbf{0.70} & \textbf{0.56} & \textbf{0.64} & 0.25 & 0.38 & 0.66 & 0.86 & \textbf{0.50} & \textbf{0.65} \\
\midrule
\multirow{15}{*}{Claude 3.7 Sonnet}
  & Baseline & 0.18 & 0.49 & 0.51 & 0.66 & 0.17 & 0.39 & 0.10 & 0.53 & 0.24 & 0.52 \\
 & History & 0.50 & 0.73 & 0.48 & 0.66 & 0.65 & 0.85 & 0.32 & 0.74 & 0.49 & 0.74 \\
\cmidrule(lr){2-12}
 & ReAct & 0.51 & 0.75 & 0.57 & 0.72 & 0.75 & 0.91 & 0.44 & 0.77 & 0.57 & 0.79 \\
 & Amem & 0.48 & 0.73 & 0.46 & 0.64 & 0.62 & 0.84 & 0.33 & 0.73 & 0.47 & 0.73 \\
\cmidrule(lr){2-12}
 & SelfRAG & 0.52 & 0.75 & 0.46 & 0.64 & 0.65 & 0.84 & 0.31 & 0.74 & 0.49 & 0.74 \\
 & Mem0 & 0.51 & 0.74 & 0.48 & 0.66 & 0.65 & 0.84 & 0.37 & 0.76 & 0.50 & 0.75 \\
\cmidrule(lr){2-12}
 & DC-Cu & 0.50 & 0.74 & 0.50 & 0.67 & 0.62 & 0.84 & 0.33 & 0.75 & 0.49 & 0.75 \\
 & DC-RS & 0.50 & 0.74 & 0.52 & 0.68 & 0.62 & 0.84 & 0.34 & 0.74 & 0.50 & 0.75 \\
 & AWM & 0.49 & 0.73 & 0.53 & 0.68 & 0.60 & 0.82 & 0.34 & 0.74 & 0.49 & 0.74 \\
\cmidrule(lr){2-12}
 & ExpRecent & 0.66 & 0.83 & 0.63 & 0.73 & 0.53 & 0.76 & 0.49 & 0.82 & 0.58 & 0.79 \\
 & ExpRAG & 0.74 & 0.89 & 0.62 & 0.72 & 0.72 & 0.89 & 0.46 & 0.76 & 0.63 & 0.82 \\
 & ReMem & \textbf{0.92} & \textbf{0.96} & \textbf{0.73} & \textbf{0.83} & \textbf{0.83} & \textbf{0.95} & \textbf{0.62} & \textbf{0.89} & \textbf{0.78} & \textbf{0.91} \\
\midrule
\multirow{15}{*}{Claude 3.5 Haiku}
 & Baseline & 0.11 & 0.33 & 0.38 & 0.52 & 0.15 & 0.32 & 0.08 & 0.37 & 0.18 & 0.39 \\
 & History & 0.28 & 0.58 & 0.38 & 0.57 & 0.18 & 0.38 & 0.12 & 0.49 & 0.24 & 0.51 \\
\cmidrule(lr){2-12}
 & ReAct & 0.24 & 0.58 & 0.35 & 0.52 & 0.32 & 0.53 & 0.16 & 0.55 & 0.27 & 0.55 \\
 & Amem & 0.24 & 0.55 & 0.37 & 0.58 & 0.17 & 0.35 & 0.12 & 0.45 & 0.23 & 0.48 \\
\cmidrule(lr){2-12}
 & SelfRAG & 0.26 & 0.58 & 0.38 & 0.59 & 0.22 & 0.37 & 0.14 & 0.49 & 0.25 & 0.51 \\
 & Mem0 & 0.27 & 0.56 & 0.37 & 0.57 & 0.17 & 0.37 & 0.08 & 0.45 & 0.22 & 0.49 \\
\cmidrule(lr){2-12}
 & DC-Cu & 0.24 & 0.55 & 0.37 & 0.58 & 0.17 & 0.37 & 0.12 & 0.45 & 0.23 & 0.49 \\
 & DC-RS & 0.24 & 0.55 & 0.37 & 0.58 & 0.17 & 0.37 & 0.12 & 0.45 & 0.23 & 0.49 \\
 & AWM & 0.24 & 0.55 & 0.37 & 0.58 & 0.17 & 0.37 & 0.12 & 0.45 & 0.23 & 0.49 \\
\cmidrule(lr){2-12}
 & ExpRecent & 0.48 & 0.65 & 0.40 & 0.57 & 0.15 & 0.32 & 0.32 & 0.64 & 0.34 & 0.55 \\
 & ExpRAG & 0.65 & 0.74 & 0.54 & 0.64 & \textbf{0.43} & \textbf{0.61} & 0.42 & 0.68 & 0.51 & 0.67 \\
 & \textbf{ReMem} & \textbf{0.69} & \textbf{0.80} & \textbf{0.49} & \textbf{0.60} & \textbf{0.43} & \textbf{0.61} & \textbf{0.44} & \textbf{0.75} & \textbf{0.51} & \textbf{0.69} \\
\bottomrule
\end{tabular}}
\caption{Cross-environment results across four embodied reasoning benchmarks (AlfWorld, BabyAI, PDDL, ScienceWorld). Each dataset reports success (S) and progress (P) rates. Bold indicates the best (including ties) per column. The last two columns show averaged S and P across datasets.}
\label{tab:multi_full}
\end{table*}

\begin{table*}[t!]
\centering
\scriptsize
\setlength{\tabcolsep}{3.5pt}
\renewcommand{\arraystretch}{1.1}
\scalebox{0.85}{
\begin{tabular}{l l*{8}{c}}
\toprule
\multirow{2}{*}{\textbf{LLM Backbone}} & \multirow{2}{*}{\textbf{Method}} & \multicolumn{6}{c}{\textbf{Exact Match $\uparrow$}} & \multicolumn{1}{c}{\textbf{API / Acc. $\uparrow$}}\\
\cmidrule(lr){3-8}\cmidrule(lr){9-9}
 &  & AIME24 & AIME25 & GPQA & MMLU-Pro (Eco.) & MMLU-Pro (Eng.) & MMLU-Pro (Philo.) & ToolBench& \multicolumn{1}{c}{\textbf{Avg. $\uparrow$}}\\
\midrule

% ---------------- Claude 3.5 ----------------
\multirow{15}{*}{Claude 3.5}
 & Baseline   & \textemdash{} & \textemdash{} & 0.36 & 0.68 & 0.42 & 0.55 & 0.81/0.64 & 0.38 \\
 & History    & \textemdash{} & \textemdash{} & 0.37 & 0.70 & 0.43 & 0.55 & 0.81/0.63 & 0.38 \\
\cmidrule(lr){2-10}
 & ReAct      & \textemdash{} & \textemdash{} & 0.35 & 0.69 & 0.43 & 0.54 & 0.81/0.64 & 0.38 \\
 & Amem       & \textemdash{} & \textemdash{} & 0.34 & 0.69 & 0.43 & 0.53 & 0.82/0.63 & 0.37 \\
\cmidrule(lr){2-10}
 & SelfRAG    & \textemdash{} & \textemdash{} & 0.36 & 0.70 & 0.44 & 0.56 & 0.83/0.65 & 0.39 \\
 & MemOS      & \textemdash{} & \textemdash{} & 0.37 & 0.70 & 0.42 & 0.55 & 0.81/0.64 & 0.38 \\
 & Mem0       & \textemdash{} & \textemdash{} & 0.36 & 0.70 & 0.42 & 0.55 & 0.82/0.64 & 0.38 \\
 & LangMem    & \textemdash{} & \textemdash{} & \textbf{0.51} & \textbf{0.78} & 0.46 & 0.61 & 0.81/0.63 & \textbf{0.43} \\
\cmidrule(lr){2-10}
 & DC-RS      & \textemdash{} & \textemdash{} & 0.36 & 0.68 & 0.41 & 0.56 & 0.82/0.64 & 0.38 \\
 & AWM        & \textemdash{} & \textemdash{} & 0.33 & 0.67 & 0.42 & 0.53 & 0.81/0.63 & 0.37 \\
 & DC-Cu      & \textemdash{} & \textemdash{} & 0.33 & 0.65 & 0.39 & 0.54 & 0.82/0.63 & 0.36 \\
\cmidrule(lr){2-10}
 & ExpRecent  & \textemdash{} & \textemdash{} & 0.42 & 0.69 & 0.46 & 0.59 & 0.85/0.63 & 0.40 \\
 & ExpRAG     & \textemdash{} & \textemdash{} & 0.40 & 0.73 & \textbf{0.49} & 0.61 & 0.87/0.67 & 0.41 \\
 & ReMem      & \textemdash{} & \textemdash{} & 0.39 & 0.71 & 0.47 & \textbf{0.62} & \textbf{0.87/0.68} & 0.41 \\
\midrule

% ---------------- Claude 3.7 ----------------
\multirow{15}{*}{Claude 3.7 Sonnet}
 & Baseline            & 0.17 & 0.13 & 0.55 & 0.84 & 0.63 & 0.78 & 0.76/0.62 & 0.54 \\
 & History       & 0.13 & \textbf{0.23} & 0.56 & 0.85 & 0.64 & 0.78 & 0.76/0.61 & 0.55 \\
\cmidrule(lr){2-10}
 & ReAct                     & 0.17 & 0.10 & 0.57 & 0.84 & 0.63 & 0.76 & 0.76/0.61 & 0.54 \\
 & Amem                      & \textbf{0.27} & 0.17 & 0.54 & 0.83 & 0.63 & 0.79 & 0.77/0.63 & 0.56 \\
\cmidrule(lr){2-10}
 & SelfRAG                   & 0.20 & 0.10 & 0.58 & 0.84 & 0.65 & 0.77 & 0.77/0.63 & 0.55 \\
 & MemOS                     & 0.17 & 0.20 & 0.55 & 0.84 & 0.64 & 0.76 & 0.76/0.62 & 0.55 \\
 & Mem0                      & 0.20 & 0.13 & 0.58 & 0.84 & 0.62 & 0.77 & 0.76/0.61 & 0.55 \\
 & LangMem                   & 0.10 & 0.13 & 0.53 & 0.77 & 0.56 & 0.66 & 0.77/0.63 & 0.49 \\
\cmidrule(lr){2-10}
 & DC-RS                     & 0.20 & 0.20 & 0.62 & 0.79 & 0.52 & 0.60 & 0.77/0.62 & 0.52 \\
 & AWM                       & 0.03 & 0.03 & 0.53 & 0.80 & 0.56 & 0.72 & 0.76/0.62 & 0.48 \\
 & DC-Cu                     & 0.17 & \textbf{0.23} & 0.57 & 0.79 & 0.52 & 0.65 & 0.77/0.62 & 0.52 \\
\cmidrule(lr){2-10}
 & ExpRecent                 & 0.13 & 0.20 & 0.61 & \textbf{0.86} & 0.63 & 0.78 & 0.82/0.66 & 0.56 \\
 & ExpRAG                    & 0.17 & 0.17 & \textbf{0.70} & 0.85 & \textbf{0.67} & \textbf{0.80} & \textbf{0.88/0.72} & \textbf{0.59} \\
 & ReMem                     & 0.13 & 0.13 & 0.67 & \textbf{0.86} & 0.65 & 0.80 & 0.87/0.71 & 0.58 \\
\midrule

% ---------------- Gemini 2.5 Flash ----------------
\multirow{15}{*}{Gemini 2.5 Flash}
 & Baseline           & 0.47 & 0.47 & 0.48 & 0.83 & \textbf{0.46} & 0.75 & 0.71/0.61 & 0.59 \\
 & History       & 0.60 & 0.47 & 0.43 & 0.84 & 0.42 & 0.78 & 0.31/0.26 & 0.55 \\
\cmidrule(lr){2-10}
 & ReAct                    & 0.30 & 0.27 & 0.05 & 0.64 & 0.16 & 0.54 & 0.64/0.57 & 0.37 \\
 & Amem                     & \textbf{0.70} & \textbf{0.57} & 0.52 & 0.83 & 0.42 & 0.72 & 0.72/0.60 & 0.63 \\
\cmidrule(lr){2-10}
 & SelfRAG                  & 0.50 & 0.47 & 0.46 & 0.83 & 0.45 & 0.75 & 0.72/0.61 & 0.59 \\
 & MemOS                    & 0.47 & 0.47 & 0.50 & 0.82 & \textbf{0.46} & 0.75 & 0.71/0.61 & 0.59 \\
 & Mem0                     & 0.50 & 0.47 & 0.45 & 0.83 & \textbf{0.46} & 0.74 & 0.71/0.61 & 0.59 \\
 & LangMem                  & 0.43 & 0.50 & \textbf{0.53} & 0.79 & 0.39 & 0.71 & 0.68/0.57 & 0.57 \\
\cmidrule(lr){2-10}
 & DC-RS                    & 0.53 & 0.37 & 0.48 & 0.80 & 0.42 & 0.69 & 0.68/0.57 & 0.56 \\
 & DC-Cu                    & 0.60 & 0.40 & 0.48 & 0.79 & 0.44 & 0.69 & 0.70/0.59 & 0.58 \\
 & AWM                      & 0.50 & 0.37 & 0.49 & 0.79 & 0.43 & 0.72 & 0.71/0.59 & 0.56 \\
\cmidrule(lr){2-10}
 & ExpRecent                & 0.47 & 0.47 & 0.42 & 0.83 & 0.39 & 0.75 & 0.78/0.66 & 0.58 \\
 & ExpRAG                   & 0.43 & 0.47 & 0.42 & 0.83 & 0.43 & 0.78 & \textbf{0.87/0.73} & 0.60 \\
 & ReMem                    & 0.60 & 0.53 & 0.51 & \textbf{0.85} & \textbf{0.46} & \textbf{0.79} & 0.85/0.71 & \textbf{0.65} \\
\midrule

% ---------------- Gemini 2.5 Flash-Lite ----------------
\multirow{15}{*}{Gemini 2.5 Flash-Lite}
 & Baseline          & 0.53 & \textbf{0.43} & 0.37 & 0.73 & 0.34 & 0.60 & 0.78/0.61 & 0.58 \\
 & History     & 0.40 & 0.33 & 0.31 & 0.74 & 0.30 & 0.59 & 0.58/0.47 & 0.49 \\
\cmidrule(lr){2-10}
 & ReAct                   & 0.53 & 0.33 & 0.34 & 0.61 & 0.20 & 0.48 & 0.73/0.56 & 0.50 \\
 & Amem                    & 0.40 & 0.33 & 0.33 & 0.72 & \textbf{0.34} & 0.59 & 0.77/0.61 & 0.54 \\
\cmidrule(lr){2-10}
 & SelfRAG                 & \textbf{0.57} & 0.37 & 0.37 & 0.73 & 0.32 & 0.62 & 0.81/0.63 & 0.57 \\
 & MemOS                   & 0.53 & \textbf{0.43} & 0.37 & 0.73 & 0.34 & 0.60 & 0.78/0.61 & 0.58 \\
 & Mem0                    & 0.53 & \textbf{0.43} & 0.37 & 0.73 & 0.34 & 0.60 & 0.78/0.61 & 0.58 \\
 & LangMem                 & 0.40 & 0.37 & \textbf{0.48} & 0.59 & 0.24 & 0.56 & 0.33/0.26 & 0.43 \\
\cmidrule(lr){2-10}
 & DC-RS                   & 0.53 & \textbf{0.43} & 0.34 & 0.59 & 0.18 & 0.36 & 0.73/0.56 & 0.49 \\
 & AWM                     & 0.03 & 0.03 & 0.35 & 0.61 & 0.20 & 0.48 & 0.77/0.60 & 0.44 \\
 & DC-Cu                   & 0.53 & \textbf{0.43} & 0.33 & 0.55 & 0.16 & 0.32 & 0.71/0.56 & 0.47 \\
\cmidrule(lr){2-10}
 & ExpRecent               & \textbf{0.57} & 0.33 & 0.35 & 0.76 & 0.29 & 0.62 & 0.82/0.65 & 0.58 \\
 & ExpRAG                  & 0.47 & 0.37 & 0.38 & \textbf{0.79} & 0.32 & \textbf{0.66} & \textbf{0.87/0.68} & \textbf{0.61} \\
 & ReMem                   & \textbf{0.57} & 0.33 & 0.38 & 0.77 & \textbf{0.34} & 0.65 & 0.86/0.67 & \textbf{0.61} \\
\bottomrule
\end{tabular}}
\caption{Cross-dataset results of diverse memory architectures across models. 
Categories are separated by horizontal rules; results (Exact Match↑ and API/Acc↑) compare zero-shot, agentic, adaptive, procedural, and proposed memory methods. 
Dashes (—) indicate methods with poor or unreliable performance, which are therefore omitted.}
\label{tab:single_full}
\end{table*}

\begin{figure}[t!]
    \centering
    \includegraphics[width=0.95\linewidth]{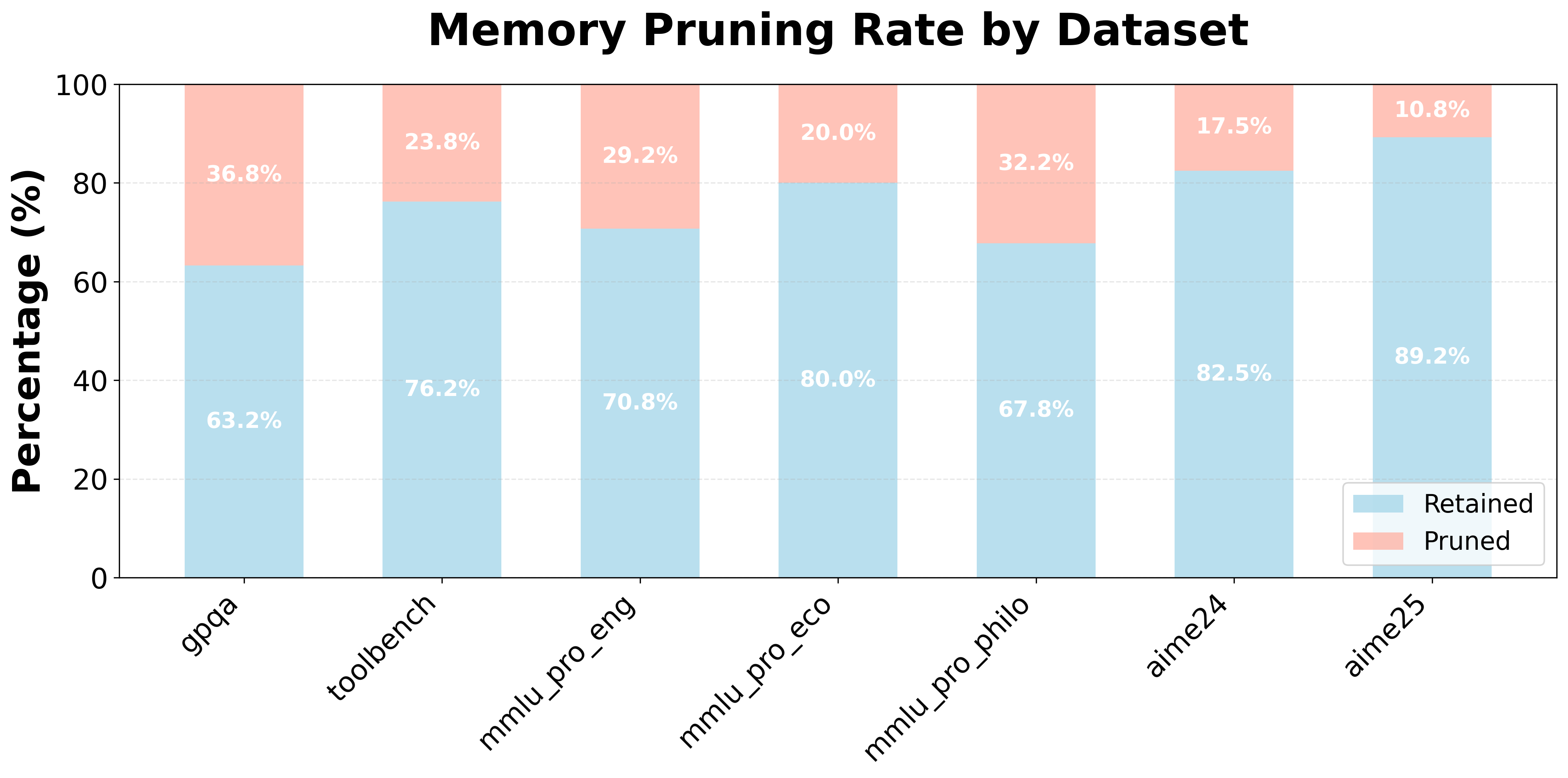}
    \caption{Memory pruning rates by dataset. Retained (blue) and pruned (coral) 
memory proportions show varying selectivity across benchmarks.}
    \label{fig:prune}
\end{figure}
\subsection{Additional Analysis of Memory Pruning}
\label{app:memory_prune}
Figure~\ref{fig:prune} shows memory pruning rates across datasets, revealing varying selectivity in memory retention. The pruning ratios differ substantially across benchmarks, which appears related to task diversity and domain coverage. Datasets with broader domain coverage such as GPQA, which encompasses diverse problem types across engineering, physics, and other domains, exhibit higher pruning rates (36.8\%), suggesting that more memories are deemed redundant across heterogeneous tasks. In contrast, datasets with more concentrated problem types like AIME show lower pruning rates (17.5\% and 10.8\% respectively), indicating that memories remain more relevant due to higher task similarity. This pattern suggests that the pruning mechanism effectively identifies and discards domain-irrelevant experiences, though the precise relationship between task diversity and memory selectivity warrants further investigation.

\subsection{Cumulative Accuracy Across Tasks and Models}
\label{app:curve}

\begin{figure}[t!]
    \centering
    \includegraphics[width=0.8\linewidth]{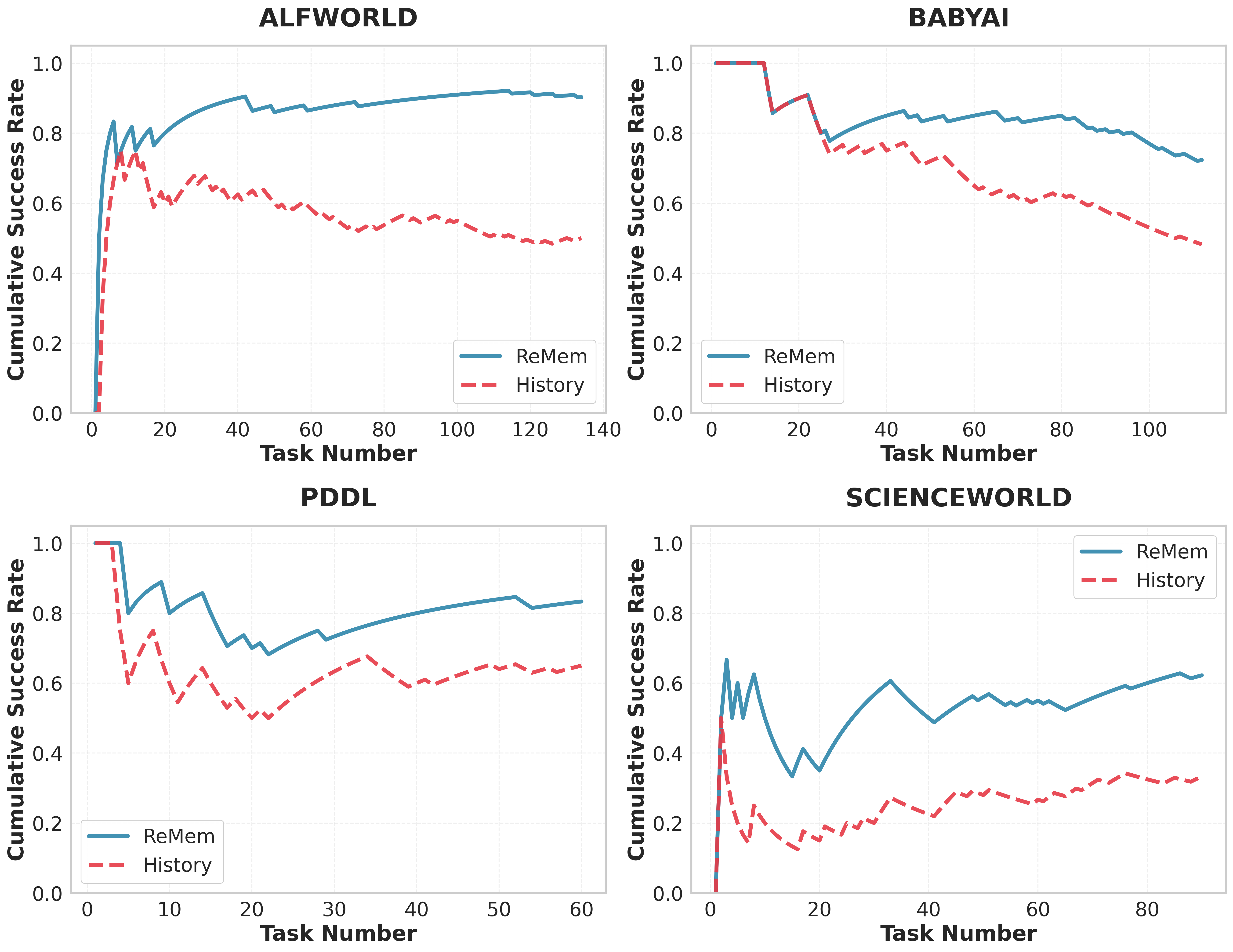}
    \caption{Cumulative success rate across four interactive agent datasets, shown as rolling averages over fixed task sequences (not learning curves).}
    \label{fig:curve}
    \vspace{-0.4cm}
\end{figure}

Figure~\ref{fig:curve} shows cumulative accuracy over task sequences across four interactive multi-turn datasets. 
The curves primarily compare \textsc{ReMem} with the History baseline, as individual trajectories are not meaningful in isolation. 
Across all environments, \textsc{ReMem} exhibits faster adaptation and more stable retention, demonstrating robustness under long task sequences. 

\begin{figure}[t!]
    \centering
    \includegraphics[width=0.95\linewidth]{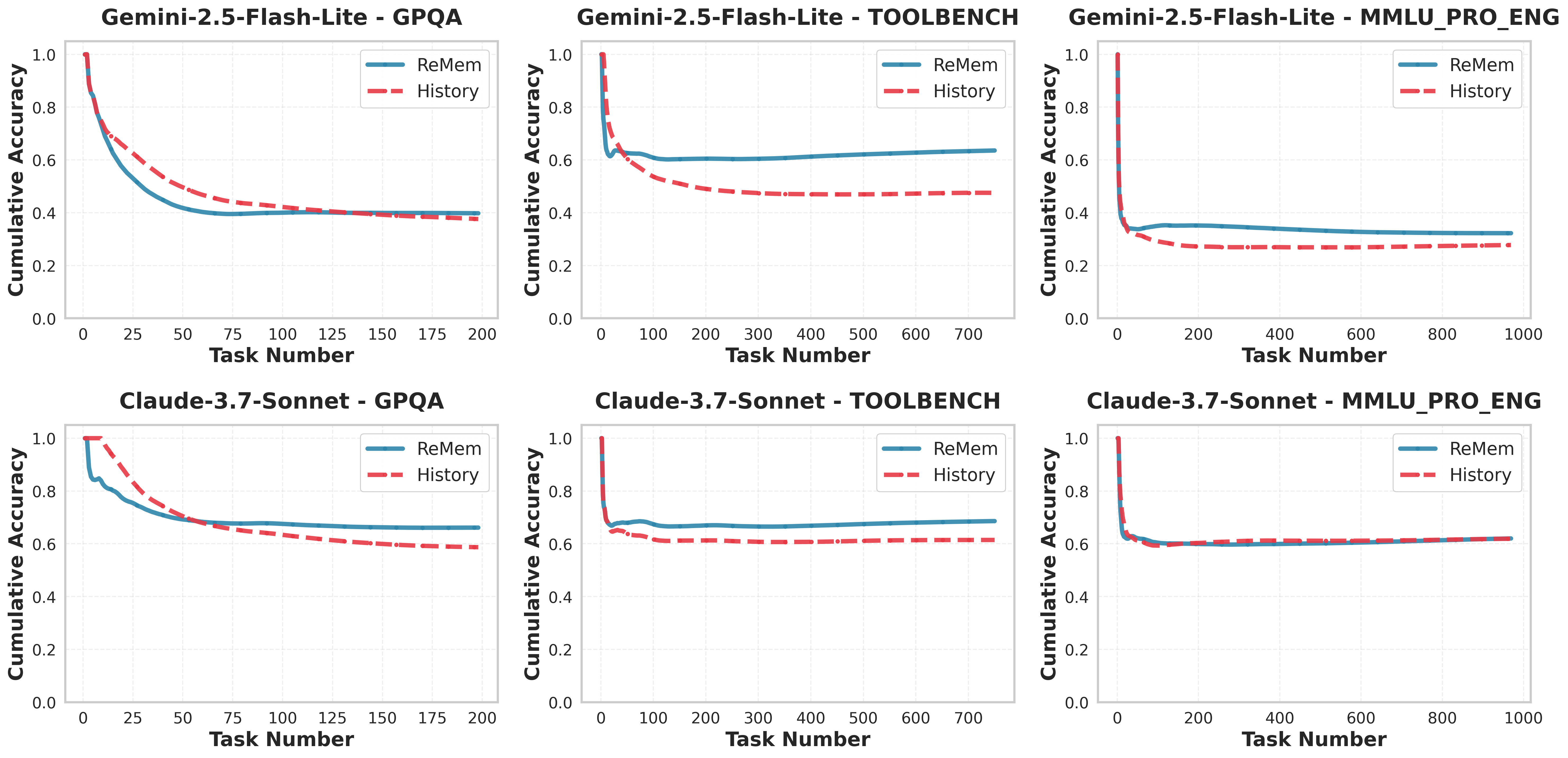}
    \caption{Cumulative accuracy comparison across model variants and benchmarks. 
ReMem (solid blue) demonstrates consistent improvements over the History baseline 
(dashed red) across Gemini-2.5-Flash-Lite and Claude-3.7-Sonnet models on GPQA, 
ToolBench, and MMLU-Pro (ENG) datasets. The curves show learning trends as task 
instances accumulate, with ReMem achieving faster convergence and higher final accuracy.}
    \label{fig:compare_single}
\end{figure}

Figure~\ref{fig:compare_single} presents cumulative accuracy curves comparing \alg with the baseline across single-turn reasoning benchmarks and model variants.
As task instances accumulate, \alg shows consistent improvement on GPQA, ToolBench, and MMLU-PRO (Engineer) for both Gemini-2.5-Flash-Lite and Claude-3.7-Sonnet.
Similar to the multi-turn results, History performs comparably at the beginning due to the cold-start phase, but \alg quickly surpasses it as more tasks are processed, indicating the cumulative advantage of continual task-level adaptation.

\section{Potential Risks}
Agent memory management relies on the outputs and judgments of large language models, and is therefore subject to their reliability limitations. 
Imperfect or inconsistent LLM outputs may lead to unreliable memory updates, which can affect subsequent retrieval and decision making. 
In addition, memory-based agents may be vulnerable to attacks or poisoning, where adversarial or misleading interactions introduce corrupted experiences into memory. 
Addressing reliability, robustness, and security concerns in LLM-driven memory management remains an important direction for future work.

\newpage
\newpage
\newpage
\newpage
\newpage
\newpage
\section{Prompts}
\label{app:prompt}
\begin{prompt}{Memory Prompt Template for Multi-turn Dataset}
\small
\textbf{=====================================}

\textbf{ENVIRONMENT INSTRUCTIONS}

\textbf{=====================================}

\textit{[Detailed task environment description and rules]}

\textit{Example: Go to kitchen, pick up apple, put it in bag, etc.}

\vspace{12pt}
\textbf{=====================================}

\textbf{EXAMPLE DEMONSTRATIONS}

\textbf{=====================================}

\textit{[Static few-shot examples]}

\textit{Example 1: Goal: ... | Action: ... | Observation: ...}

\textit{Example 2: Goal: ... | Action: ... | Observation: ...}

\vspace{12pt}
\textbf{=====================================}

\textbf{RELEVANT EXPERIENCE FROM SIMILAR TASKS}

\textbf{=====================================}

\textit{[Experience \#1]}

\textit{Goal: [similar goal]}

\textit{Trajectory: [action sequence]}

\textit{Correctness: [success/failure]}

\vspace{6pt}
\textit{[Experience \#2, \#3, ...]}

\vspace{12pt}
\textbf{=====================================}

\textbf{YOUR CURRENT TASK}

\textbf{=====================================}

\textbf{Goal:} \textit{[specific task goal]}

\textit{Help: type 'check valid actions' if action fails}

\textit{Help: type 'inventory' to check items}

\vspace{12pt}
\textbf{=====================================}

\textbf{RECENT HISTORY}

\textbf{=====================================}

\texttt{Observation: [initial environment state]}

\texttt{Action: [previous action]}

\texttt{Observation: [result of previous action]}

\texttt{Action: [previous action]}

\texttt{Observation: [current state]}

\vspace{12pt}
\textbf{=====================================}

\textbf{OUTPUT FORMAT}

\textbf{=====================================}

You MUST respond in EXACTLY ONE of these formats:

\vspace{8pt}
\textbf{Format 1 - Prune experiences:}

\texttt{Think-Prune: <IDs>}

Remove unhelpful experiences from 'RELEVANT EXPERIENCE' section (e.g., ``1,3'' or ``2-4'')

\vspace{8pt}
\textbf{Format 2 - Internal reasoning:}

\texttt{Think: <your reasoning>}

Free-form explanation of your next step

\vspace{8pt}
\textbf{Format 3 - Execute action:}

\texttt{Action: <exact command>}

Must be valid command from ENVIRONMENT INSTRUCTIONS with exact names from RECENT HISTORY

\end{prompt}

\begin{prompt}{Memory Prompt Template for Single-turn Dataset}
You are a helpful assistant with access to LOCAL EXPERIENCE MEMORY. Each memory may contain past experience, rationales, domains, and skills. Below are some retrieved LOCAL EXPERIENCE MEMORIES:

\vspace{6pt}
\textit{[Retrieved/synthesized memories]}

\vspace{6pt}
Now solve the following problem.

\textbf{Question:} \textit{[Your question here]}

\vspace{6pt}
\noindent\textbf{Provide your output in the following format:}
\begin{itemize}
  \item \textbf{Rationale:} your short reasoning, may cite memory if useful
  \item \textbf{Final Answer:} your final answer
\end{itemize}
\end{prompt}

\section{Ethical Considerations and Artifact Documentation}

\subsection{Cite Creators of Artifacts}
All external artifacts used in this work are properly credited to their original publications and repositories in Section \ref{exp_setup}.

\subsection{Discuss the License for Artifacts}
We comply with the licenses and terms of use of all artifacts employed in this study. The benchmark datasets, such as MMLU-Pro, GPQA, AlfWorld, etc., are publicly available for research and evaluation purposes under their respective licenses (typically Creative Commons or similar non-commercial research licenses). Access to proprietary LLMs, such as Gemini and Claude, is conducted through official APIs in accordance with the providers’ usage policies.

All newly introduced code, configurations, and evaluation pipelines for Evo-Memory will be released under the permissive open-source license to support transparency and reproducibility upon acceptance.

\subsection{Artifact Use Consistent with Intended Purpose}
We confirm that all datasets and models are used in accordance with their intended purposes. The selected benchmarks are designed to evaluate reasoning, factual recall, tool use, and multi-turn interaction, which directly aligns with our goal of assessing test-time learning and self-evolving memory in LLM agents.

The evaluated LLMs are used solely for inference and controlled test-time interaction, without modifying their pre-trained parameters. Memory modules operate externally and do not alter the underlying model weights, ensuring consistency with the original model design.

\subsection{Personally Identifying Information or Offensive Content}
The datasets used in this work are established academic benchmarks widely adopted by the research community. To the best of our knowledge, they do not contain personally identifying information. While some datasets may reflect biases or problematic content inherited from web-scale sources, these issues are not unique to our work and are consistent with known limitations of the original benchmarks.

We do not introduce new user-generated content or collect personal data as part of this study.

\subsection{Documentation of Artifacts}
We provide detailed documentation of all datasets, model backbones, and memory methods evaluated in this work. Appendix \ref{app:method} and \ref{app:config} details the compared memory mechanisms and implementation choices. Prompt templates and configuration files are included in Appendix \ref{app:prompt} to facilitate full reproducibility.

\section{Statistics for Data}
Dataset statistics are summarized below. Additional details are provided in Appendix \ref{app:dataset}.

\subsection{Single-Turn Reasoning Datasets}
\begin{itemize}[noitemsep, topsep=0pt]
  \item \textbf{MMLU-Pro}: A multi-domain multiple-choice benchmark consisting of approximately 12K questions spanning 14 subject areas, including engineering, economics, philosophy, and natural sciences. 
  In our evaluation, we include major domains such as Engineering (969 questions, aggregating subfields such as EE and ME), Economics (844 questions, covering both macro- and micro-economics), and Philosophy (499 questions).
  
  \item \textbf{GPQA-Diamond}: 198 expert-curated, graduate-level multiple-choice questions designed to be Google-proof, focusing on deep scientific reasoning.
  
  \item \textbf{AIME-24 / AIME-25}: 30 Olympiad-style mathematics problems per year, requiring exact-match symbolic reasoning.
  
  \item \textbf{ToolBench}: 750 tool-use and API grounding tasks, evaluating the model’s ability to select and invoke appropriate tools, measured by exact match and execution-based accuracy.
\end{itemize}

\subsection{Multi-Turn Interactive Environments}
\begin{itemize}[noitemsep, topsep=0pt]
  \item \textbf{AlfWorld}: 134 text-based embodied household tasks, requiring multi-step planning and interaction.
  
  \item \textbf{BabyAI}: 112 grid-based navigation and object manipulation tasks with compositional language instructions.
  
  \item \textbf{ScienceWorld}: 90 interactive science tasks spanning physics, chemistry, and biology, requiring long-horizon reasoning.
  
  \item \textbf{PDDL}: 60 symbolic planning problems expressed, evaluating goal-directed planning and state transitions.
\end{itemize}

\section{Computational Experiments}
All experiments are fully reproducible, with implementation details provided in Appendix \ref{appendix:setup} and prompts listed in Appendix \ref{app:prompt}.

\subsection{Models and Budgets}
We evaluate Evo-Memory on multiple LLM backbones with varying capacities, including Gemini-2.5 Flash, Flash-Lite, and Pro, as well as Claude 3.5 Haiku and Claude 3.7 Sonnet. All experiments are conducted using fixed API budgets across methods to ensure fair comparison. The total API compute cost was on the order of tens of thousands of US dollars.

\subsection{Descriptive Statistics}
Performance is reported using task-appropriate metrics, including exact match for single-turn reasoning tasks, success and progress rates for multi-turn environments, as well as cumulative accuracy and robustness over task streams. Reported results are averaged across multiple task instances and trajectories.

\subsection{Parameters for Packages}
All experiments are implemented using standard Python tooling. API-based access to LLMs is handled through the official SDKs provided by model vendors. Additional utilities for retrieval, logging, and evaluation are implemented using PyTorch 2.7.1 and Weights \& Biases.

\section{AI Assistants in Research or Writing}
During the preparation of this paper, we made limited and controlled use of large language models (LLMs), specifically ChatGPT, as an auxiliary writing aid. 
The LLM was used only for stylistic refinement, including improvements in clarity, grammar, and readability of text originally drafted by the authors. 
All scientific ideas, analyses, experiments, and conclusions were fully developed, written, and verified by the authors. 
Thus, LLMs were employed solely as a language-editing tool, without contributing to the intellectual or scientific content of the work.

\section{Limitations}
While \method offers a comprehensive evaluation of self-evolving memory, several practical constraints remain. 
Due to budget and API limits, we focus on a selected set of strong LLMs rather than exhaustively covering all available models. 
Additional evaluations on open-weight or multilingual models could further validate the generality of our findings. 
Moreover, our benchmark primarily emphasizes textual and goal-oriented tasks; extending it to richer multimodal or real-world environments would provide a more complete picture of continual memory evolution. 
Despite these limitations, the current study already spans diverse domains, tasks, and architectures, offering a solid foundation for future extensions.


\end{document}